\documentclass{report}

\usepackage[preprint]{nips}
\usepackage{multirow}
\usepackage{CJKutf8}
\usepackage{graphicx} 
\usepackage{makecell} 
\usepackage{booktabs} 
\usepackage{float} 

\usepackage[utf8]{inputenc} 
\usepackage[T1]{fontenc}    
\usepackage{hyperref}       
\usepackage{url}            
\usepackage{booktabs}       
\usepackage{amsfonts}       
\usepackage{nicefrac}       
\usepackage{microtype}      
\usepackage{xcolor}         
\usepackage{parskip} 
\usepackage{hyperref} 
\usepackage{parskip} 
\usepackage{enumitem} 
\usepackage{hyperref} 
\usepackage{authblk}
\newcommand{\Background}[1]{
    \par\bigskip 
    \noindent{\Large\textbf{I. #1}} 
    \par\medskip 
}
\newcommand{\hmit}[1]{
    \par\bigskip 
    \noindent{\Large\textbf{II. #1}} 
    \par\medskip 
}
\newcommand{\ontribution}[1]{
    \par\bigskip 
    \noindent{\Large\textbf{III. #1}} 
    \par\medskip 
}
\newcommand{\Impact}[1]{
    \par\bigskip 
    \noindent{\Large\textbf{IV. #1}} 
    \par\medskip 
}
\newcommand{\way}[1]{
    \par\bigskip 
    \noindent{\Large\textbf{V. #1}} 
    \par\medskip 
}

\definecolor{lightblue}{RGB}{100, 100, 255} 
\definecolor{lightred}{RGB}{255, 100, 100} 

\begin{document}


\title{\fontsize{15pt}{30pt}\selectfont \setlength{\linewidth}{\textwidth}%
Solving the Unsolvable: Translating Case Law in Hong Kong}


%



\author{%
  \textbf{King-kui Sin}$^1$\thanks{Corresponding author}, \textbf{Xi Xuan}$^2$, \textbf{Chunyu Kit}$^2$\thanks{Corresponding author}, \textbf{Clara Ho-yan Chan}$^3$, \textbf{Honic Ho-kin Ip}$^4$ \\
  $^1$UOW College Hong Kong, $^2$City University of Hong Kong, \\
  $^3$The Chinese University of Hong Kong, Shenzhen, 
  $^4$The University of Hong Kong SPACE
}

\maketitle

\begin{abstract}

This paper addresses the challenges translating case law under Hong Kong's bilingual legal system. It highlights the initial success of translating all written statutes into Chinese before the 1997 handover, a task mandated by the Basic Law. The effort involved significant collaboration among legal, linguistic, and translation experts, resulting in a comprehensive and culturally appropriate bilingual legal system. However, translating case law remains a significant challenge due to the sheer volume and continuous growth of judicial decisions. The paper critiques the government’s and judiciary’s sporadic and uncoordinated efforts to translate case law, contrasting it with the thorough approach previously taken for statute translation. Although the government acknowledges the importance of legal bilingualism, it lacks a sustainable strategy for translating case law. The Judiciary’s position—that translating all judgments is “unnecessary, unrealistic, and not cost-effective”—is analyzed and critiqued for its impact on legal transparency and public trust. A proposed solution involves leveraging machine translation technology through a human-machine interactive translation platform, which undergoes two major transitions. Initially based on a neural model, the platform transitions to using a large language model for improved translation accuracy. Furthermore, it evolves from a single-agent system to a multi-agent system, incorporating Translator, Annotator, and Proofreader agents. This multi-agent approach, supported by a grant, aims to facilitate efficient, high-quality translation of judicial judgments by integrating advanced artificial intelligence and continuous feedback mechanisms, thus better meeting the needs of a bilingual legal system.

\end{abstract}

\Background{Background}

\section{Bilingual legislation}

This year marks the 27\textsuperscript{th} anniversary of Hong Kong’s return to China after 155 years of British rule. On 1 July 1997, when the Hong Kong Special Administrative Region (HKSAR) was established, it took pride in having all its written statutes translated into Chinese, a mammoth task which had been regarded by many as impossible. It was a task dictated by Articles 8 and 9 of the \textit{Basic Law of Hong Kong}, the former stipulating the retention of the English common law system in the HKSAR, and the latter legal bilingualism, i.e., a common law system operating in Chinese
as the default language of the law, with the optional use of English as an official language.\setcounter{footnote}{0}
\footnote{Article 9 provides: “In addition to the Chinese language, English may also be used as an official language by the executive authorities, legislature and judiciary of the Hong Kong Administrative Region.” While Chinese and English are both official languages under Section 3 of the \textit{Official Languages Ordinance} (Cap 5 of the Laws of Hong Kong), the wording suggests a subtle distinction in pragmatic meaning between the two. As part of China, it goes without saying that Hong Kong’s legal system must operate in the language of the sovereign state, namely, Chinese. It is therefore only natural that Chinese is the \textit{default} language of the law, as implied by the prepositional phrase “in addition to Chinese.” As for English, since it is an official language, it may be so used, but its use is only \textit{optional}. 

In this connection, Lee (2020) has misread the pragmatic meaning of “in addition to” in Article 9, arguing that “the wording bears the subtle suggestion that the Chinese language was always already an official language used by the three major government institutions, to be supplemented by English, which \textit{may also} ... be used for similar purposes” (Lee 2020, p.867). Promulgated in 1990, Article 9 of the \textit{Basic Law} stipulated what was to happen in respect of the languages of the law in Hong Kong after 1997, not before. Nevertheless, he is right in noting that the use of English is optional. This also reinforces the default status of Chinese in the HKSAR.}

To prepare Hong Kong for the transformation from a monolingual to a bilingual legal system in the
run-up to 1997, the then Legal Department (now the Department of Justice) took proactive
measures to implement bilingual legislation. Soon after the announcement of the \textit{Sino-British
Joint Declaration} in 1984, it set up a working party in 1985 to launch the Bilingual Laws Project.
In 1987 it amended the \textit{Interpretation and General Clauses Ordinance} to give equal legal status
to the Chinese text of the law alongside its English counterpart. In the same year it amended the
\textit{official Languages Ordinance}, requiring that subsequent legislation be enacted in both Chinese
and English whereby legal bilingualism was formally established. In 1988 the Bilingual Laws
Advisory Committee (BLAC) was established under the amended \textit{Official Language Ordinance}
to advise the Government on the translation of legislation. In 1989 translation of legislation
formally commenced. And in May 1997, two months before the handover, the task was completed
– all 514 existing ordinances, approximately 10 million words on 21,000 pages, were translated.\footnote{For a detailed account of the translation process, see Sin (1999). And for a more detailed account of Hong Kong’s
journey to legal bilingualism before 1997, see Zhao (1997).}

Of all former British colonies, Hong Kong is so far the only one which succeeded in having its
entire statutory law available in the local language on the very first day when it ceased to be under
British rule.

\section{Chinese digest}

However, having all the written statutes available in Chinese and English is by no means a
full implementation of legal bilingualism. For written statutes represent only a small portion of
the law of Hong Kong. The bulk is found in judicial judgments of cases decided by the superior
courts of Hong Kong and other common law jurisdictions, known as case law. Such judgments
are voluminous, and what is more, ever growing in number as cases are being tried in common
law jurisdictions across the world every day. While only a tiny portion of such judgments are
written in Chinese – they are all judgments of local cases decided in Hong Kong, an
overwhelmingly larger portion are written in English. Thus, implementing legal bilingualism in
Hong Kong requires translating all translating all English judgments into Chinese, and vice versa. 

Given the comparatively small number of Chinese judgments, translating all of them into English does not seem an insurmountable task. In contrast, given the immensely greater number of English judgments, translating all of them into Chinese is apparently an impossible task.

Way back in 1984, when China first indicated its intention to retain the common law system in the HKSAR as a bilingual legal system operating in Chinese and English, many legal experts already noticed the problem inherent in this policy. Henry Litton, who subsequently became a Permanent Judge of the Court of Final Appeal, pointed out flatly, “There seems to be broad consensus in Hong Kong that the translation of the whole of the common law into Chinese is impossible, common law being that broad body of law derived from decisions of the courts in England, embodied in the law reports, and fed by sources as diverse as Australia, Canada and New Zealand” \cite{litton1998}. And 38 years later, on 1 July 2022, he stressed the same point again, remarking that “there is no conceivable way that the vast bulk of cases constituting the common law could be translated into Chinese” \cite{litton2022}.

The first attempt to address the issue was taken up by a team of legal scholars and linguists in the run-up to 1997 before the handover. To get around the enormous amount of judgments that need to be translated, they employed “the well-tried traditional technique of creating a digest of the common law” \cite{roebuck1996}. The idea was to extract from relevant authorities the legal principle underlying judicial decisions on a particular issue and create a re-statement of the legal principle in question \cite{roebuck2010}. Three bilingual digests were produced (contract law: \cite{roebuck1995}; criminal law: \cite{roebuck1996}; criminal procedure: \cite{roebuck1997}). A huge number of decided cases concerning a particular domain, say, contract, are condensed into a small number of legal principles in Chinese and English, thus giving hope to an otherwise hopeless task.

However, such an attempt turned out to be more an academic exercise than a viable solution to the problem as it would create even more problems than straightforward translation. For one thing, what legal principle underlies a judicial decision is open to interpretation. Views can be drastically divided on what precisely the law is for the question at issue. Even if an agreed re-statement is arrived at, for the re-statement to become law, it has to be re-phrased in the legislative style and enacted through legislative measures, a process known as codification. Codification may take various forms, typically that of the Civil Law systems like the French Civil Code and the German Civil Code, which lay down comprehensive, systematic, general, and gap-free legal rules \cite{svantesson2008}. Alternatively, as has been suggested by some common law jurists, it may simply be “a systematic collection or formulation of the law, reducing it from a disparate mass into an accessible statement which is given legislative rather than merely judicial or academic authority” \cite{donald1973}. Since as early as the 16\textsuperscript{th} century, the call for codifying the common law has been prompted by the need to tidy up the ever-growing bulk of scattered, and at times conflicting, case law \cite{baker2007}. Attempts have been made in England and in other common law jurisdictions, yet without much success. Technical issues aside, the slow progress of codification has been due to strong resistance among common law jurists who contend that codification is in essence incompatible with the very nature of the common law system.\footnote{Joel Prentiss Bishop was one of those die-hard opponents who cried out for the stop of codification. He said, “Our
common law is a particular system of reason. It is one of the great departments of our governmental structure; and
the study, practice, and administration of it produce that training of the reason necessary to the carrying on of the
government in the other departments. And if we convert our common law of reason into statutes, we in effect abolish
reason in things legal and governmental, so that our whole system of government and law becomes a wreck. Such is
the end of a total codification” (1888, p.3).} A highly political issue dividing common lawyers into conservatives and progressives, codification was never put on the agenda of the Hong Kong Government’s Bilingual Laws Project, nor, given the gravity of the problem, would it ever be considered as a possible solution to the issue in question.


\section{Government attempts}
\label{gen_inst}
While focusing its efforts on the translation of the written statutes before the handover, the Government did not overlook the need for the translation of the case law. In January 1997, it announced its plan for “an overall co-ordination mechanism for overseeing the implementation of a bilingual legal system in Hong Kong,” which included an item on common law cases.\footnote{The Information Paper presented to the Legislative Council on 13 January 1997 revealed the Government’s ambitious and comprehensive plan for implementing a bilingual legal system in Hong Kong. In conjunction with the Judiciary, it considered two types of issues, namely, policy issues and practical issues. Top on the policy list was “the ultimate objective of bilingualism and the question of whether Chinese and English will both be used or whether the former will eventually replace the latter.” Common law cases came under practical issues. What precisely the Government was planning to do about common law cases is not clear, as the issue was vaguely phrased as “how to relate to common law cases.” But as the Government already thought of the possibility of eventually replacing English with Chinese in law, having the case law available in Chinese must have crossed its mind. That is to say, something must be done to pave way for the materialization of that possibility.}

The next year on 1 April 1998, it established the Committee on Bilingual System (CBLS) to coordinate efforts in implementing legal bilingualism undertaken by various public and professional bodies. And recognizing the importance of having the case law available in Chinese, it established a Subcommittee on the Translation of Case Precedents in December 1998 with the objective to produce translations of judgments for citation purposes. It was also responsible for identifying judgments for translation on a pilot basis, formulating strategies for the systematic translation of case law, and devising a mechanism for vetting translations by legal professionals \cite{ng2000}. Initially, 33 cases were recommended for translation by the Judiciary, the Bar Association, the Law Society, and the Department of Justice. Twenty-five were translated variously by the Judiciary, the Official Languages Agency, and the Department of Justice. The translation was completed within a short period of three months, followed by a report in which the Subcommittee discussed the problems encountered in the process and, more importantly, announced its long-term objective to develop a sufficient body of bilingual case law \cite{cbls1999}.

Strangely enough, the Subcommittee seemed to disappear without a trace, there being no further information about its alleged commitments. But the translation of judgments remained a main concern of the lawmakers. In 2002, they raised three questions about the practice in overseas jurisdictions: First, whether all court judgments are translated; second, the legal status of court judgments in both the original and translated versions; and third, whether the translation of court judgments is undertaken by the judiciaries concerned or private agencies (LC Paper No. CB(2) 2566/02-03(01)). In response, the Department of Justice only provided some general information about the practice in Canada and Macau (Ibid). It was the Judiciary that addressed the questions directly. First, it stated unambiguously that it was never its policy that “all English judgments should be translated into Chinese, or vice versa,” as it considered this \textit{“unnecessary, unrealistic and not cost-effective”} (LC Paper No. CB (2) 1856/02-03(01); italics added). Second, it stated, also unambiguously, that the translation of a judgment has no legal status; only the original judgment is the authentic version (Ibid.). Third, it announced that the translation of whole judgments in English or Chinese was undertaken by its in-house translators of the Court Language Section, whereas “the Chinese translation of excerpts from commonly cited judgments in English of courts in Hong Kong and courts in other common law jurisdictions” would be outsourced to a legal publisher in the form of Case Books.\footnote{Three such extracts were published by Sweet \& Maxwell: one from criminal cases (2003), one from land cases (2005), and one from employment cases (2006). In this connection, it is worth noting that Sweet \& Maxwell also published a series entitled \textit{Chinese Law Reports and Translations} (1995–2008), which contains 250 original judgments in Chinese with English translations. The Chinese judgments were selected by judges in light of their jurisprudential value. The English translations were partly undertaken by the Judiciary’s Court Language Section and partly by practicing lawyers vetted by the publisher’s legal editors. Again, for reasons unknown to the public, the project came to an abrupt end.} Forensic technologies, including Automatic Speaker Verification (ASV) \cite{xuan2021research2,ding2021crowd,xuan2022research1,xuan2022research2,xuan2022multi,xuan2024conformer}, Speech Deepfake Detection (SDD) \cite{xuan2025fakemamba,xuan2025wavespnet}, and Source Tracing (ST) \cite{xuan2025multilingual} systems, are used in forensic and judicial settings but also face challenges in Hong Kong’s bilingual legal environment.

\section{The present situation}

The initial success of translating all written statutes into Chinese before the handover in 1997 was a monumental task. This accomplishment demonstrates significant dedication and organizational capability on the part of the Hong Kong Government, in particular, the Legal Department (now the Department of Justice), fulfilling a requirement stipulated by the \textit{Basic Law}. It is important to note that the translation project had significant input from legal, language, and translation scholars through the diligent work of the Bilingual Laws Advisory Committee (BLAC), a statutory body established by the Hong Kong Government to advise it on all aspects of the translation of written statutes into Chinese.

In the process of translation, a comprehensive range of issues relating to law, language, culture, and translation were revealed and explored. This extensive and collaborative effort underscores that translating laws from one language into another within a bilingual or multilingual legal system must be viewed as a social project that harnesses the full spectrum of relevant expertise in society. The successful implementation of bilingual legislation in Hong Kong is a testament to the coordinated efforts of various professionals and scholars, reflecting a commitment to accuracy, cultural sensitivity, and legal integrity.

Such a comprehensive approach ensures that the translated laws are not only linguistically accurate but also culturally appropriate and legally sound. It demonstrates that the establishment of a bilingual legal system is more than a mere administrative task; it is a complex, multifaceted undertaking that requires the integration of knowledge and skills from diverse fields. This holistic methodology is essential for preserving the principles of justice and ensuring the law is accessible and comprehensible to all members of society, regardless of their linguistic background.

However, this strong commitment to translating written statutes stands in sharp contrast to the lackluster attitude of the Government and Judiciary towards translating case law. Despite recognizing the need for legal bilingualism, the efforts to translate case law have been sporadic and uncoordinated. The establishment of the Subcommittee on the Translation of Case Precedents in 1998, which initially showed promise, eventually faded without substantial progress or a long-term strategy.

There is a notable lack of a comprehensive, sustainable strategy for translating case law, which is crucial for true legal bilingualism. While the translation of written statutes was seen as a necessary and achievable goal, the ongoing translation of case law appears to be viewed as an insurmountable challenge, leading to a lack of sustained effort and commitment. Furthermore, the inconsistent implementation, marked by the discontinuation of certain committees and unclear follow-through on initial plans, suggests significant inconsistencies in how legal bilingualism in case law is being handled.

The Judiciary’s explicit statement \cite{lc2003} that it does not intend to translate all English judgments into Chinese and vice versa, considering it “unnecessary, unrealistic, and not cost-effective,” further highlights the indifferent approach. This is compounded by the fact that translated judgments are not granted legal status, which undermines the goal of a truly bilingual legal system.

This stark contrast between the comprehensive and methodical approach taken towards statute translation and the fragmented and indifferent attitude towards case law translation reveals a significant gap in the implementation of legal bilingualism. For true legal bilingualism to be realized, a concerted effort is required not only in translating statutes but also in systematically translating case law. This requires a strategic, well-resourced, and coordinated approach that reflects the same level of dedication and collaboration demonstrated in the initial translation of written statutes. Only then can the principles of justice and accessibility be fully upheld in a bilingual legal system.

In the absence of a Chinese version of English precedents, citation of case law in court trials conducted in Chinese has been a typical sociolinguistic drama of code-switching and code-mixing. A precedent is cited in English, followed by elucidation and argument in Chinese mixed with English or in English mixed with Chinese. On very rare occasions, a bold-enough counsel may cite an English case in his/her own Chinese translation, but fearing that the court and the other party may not understand what he/she has said, cites the case again in its original English version. The admirable endeavor to use Chinese in citing case law in a Chinese trial does not seem to play any significant role in this court drama.

While such ad hoc arrangements can serve the pragmatic purposes of a court trial, if allowed to persist, it will spell the stagnation of the development of legal bilingualism. Consequently, the public’s confidence in the legal system may be eroded as a good part of the law is inaccessible in their native language, which can create significant barriers to justice.

\section{Proposed solution}

While adamant about its stated policy on the translation of judgments, the Judiciary never explains why providing a bilingual version of judgments under an avowed bilingual legal system is “unnecessary, unrealistic, and not cost-effective.” So let us conjecture what argument could be put forward for each of these claims.

First, the argument for the “unnecessary” claim may run like this: Just as not every judgment is worth reporting due to a lack of jurisprudential value, so \textit{ipso facto} not every judgment is worth translating. This argument is flawed for two reasons. First, even if we accept that not every judgment has jurisprudential value, it is difficult to pinpoint a particular judgment and say with certainty that it has no jurisprudential value whatsoever. In fact, there are judgments that initially seemed to lack significant jurisprudential value but later proved to be important. A good case in point is \textit{Thompson v Cremin}, which was originally decided in 1941 and escaped all notice until 1952 when Heuston discovered it in the Lords Journals “by good fortune alone.” Therefore, just as one cannot make the “unnecessary” claim for reporting on that ground, one cannot make it for translating on that ground either. Moreover, if jurisprudential value is what one seeks in a judgment, the process of translating, which involves scrutinizing the original text, can increase the chances of uncovering any jurisprudential value that might not have been immediately apparent. Thus, a stronger case can be made for translation. However, focusing on jurisprudential value is simply a red herring. We must not lose sight of the key role translation plays in our bilingual legal system: ensuring that all citizens, regardless of their preferred official language, have equal access to legal decisions. This is crucial for maintaining transparency and public trust in the legal system. Failing to provide translations of court rulings with significant social impacts has led to severe criticism of the Judiciary for rendering Hong Kong’s legal system inadequately transparent.\footnote{The \textit{South China Morning Post} lashed out at the Judiciary in its editorial on 14 March 2022 for “rendering Hong Kong’s system of open justice being lost in translation.” It emphasized the importance of Hong Kong's bilingual legal system, using both English and Chinese, for accessibility and reflecting the city's legal culture. The editorial noted that significant court judgments, especially in politically sensitive cases, often lack timely English translations, including cases under the security law. In a follow-up report on 17 March 2022, the newspaper reiterated the benefits of the bilingual system and stressed the necessity of translations to ensure judgments are widely understood.} The general public has become increasingly aware that to uphold open justice in a bilingual legal system, timely translation is essential.

Presumably, the “unrealistic and not cost-effective” claim is prompted by the sheer volume of judgments. When the Judiciary declared in 2003 that “[i]t has never been the policy of the Judiciary that all English judgments should be translated into Chinese, or vice versa” \cite{lc2003}, we are not sure what is intended to mean by this negation of “all.” But whatever it means, Chinese judgments are comparatively small in number as Hong Kong is the only common law jurisdiction operating in Chinese. Thus, translating all Chinese judgments into English is not an insurmountable task. The real challenge lies in the significantly larger number of English judgments in Hong Kong, not to mention the vast number of judgments in other common law jurisdictions worldwide accumulated since the inception of case law. Given China’s historical achievement of translating the entire body of Buddhist scriptures over more than a thousand years since the East Han Dynasty (57–75 AD), the idea of translating the entire body of case law may not seem so far-fetched. Moreover, with the advent of machine translation (MT) technology, mere quantity is no longer a hurdle, and we no longer need a thousand years to accomplish this task.

Puzzling as it may seem, leveraging the ever-developing technology appears to have never crossed the minds of the Judiciary. When the Judiciary declared its translation policy in 2003, machine translation (MT) was not yet evolving from the statistical model to the neural model. The latter turned out to outperform the former tremendously by understanding entire sentences rather than just piecing together small chunks of text. It uses advanced neural networks that mimic the human brain, allowing it to grasp context and nuances, making translations sound more natural and fluent. 

While the statistical model relies on matching patterns and probabilities from large datasets, it often misses the bigger picture, resulting in awkward translations. The neural model, on the other hand, learns from the context of the entire sentence, making smarter and more accurate translation choices. This huge leap in understanding and processing language context allows neural models to deliver significantly better translations. One of the most notable early advancements was Google Translate's launch of its neural machine translation (NMT) in 2016, which significantly improved translation quality and user interaction.

Human-machine interactive translation (HMIT) is intended to combine human input and machine translation to refine and enhance the translation process and quality, leveraging the strengths of both human translators and advanced machine learning models. There have long been various forms of HMIT involving different levels of interaction, often referred to as computer-assisted translation (CAT) or translation management systems (TMS). These are now commonly used to combine the efficiency of machine translation with the expertise of human translators. Genuine HMIT to minimize human effort is yet to be integrated into existing CAT platforms, allowing human translators to maximize not only their productivity, quality, creativity, and efficiency in translation but also their comfort in routine work, including:

\begin{itemize}[left=0.2in]
    \item \textbf{Review and Edit:} Quickly review machine-generated translations and make necessary corrections, improving accuracy and fluency.
    \item \textbf{Leverage Translation Memory:} Use previous translations stored in a database to ensure consistency and save time on repetitive tasks.
    \item \textbf{Post-Edit Machine Translation (PEMT):} Efficiently edit machine translations to meet quality standards required for professional use.
\end{itemize}

By integrating MT to enhance productivity, the most distinctive feature of an HMIT platform is the high speed of handling large volumes of translation work quickly and efficiently\cite{xuan2024efficient}, overcoming the challenge of sheer volume that used to overwhelm human translators. MT can process and translate huge amounts of text in a fraction of the time it would take a human, thanks to advanced algorithms\cite{xuan2025translaw, lin2025primek,xuan2021research1,zhang2025frect,zhang2025addressing,zhang2025feature}, powerful computing resources\cite{cui2025unlocking}, and, more importantly, massive volumes of bilingual data for training.\footnote{For a detailed account of the development of neural translation, see P. Koehn (2020).}

Realizing that the solution to the translation of case law lies in the exploitation of machine translation (MT) technology, we submitted a proposal to the Research Grant Council in 2020 for developing a Human-Machine Interactive Translation (HMIT) platform. This platform would utilize the latest neural machine translation (NMT) technology to facilitate efficient machine learning, user-friendly post-editing, and speedy, massive, and quality translation of judicial judgments. Our proposal was supported with a grant of about HK\$1.2 million (UGC-FDS51-H03-20).

\section{Development of HK Bilingual CFA Judgment Corpus}

However powerful an automated translation engine, it is incapable of producing good translations without good training data. The training data for an automated translation engine comprise the initial training dataset, which forms the foundational knowledge base, and the post-edited translations, which are integrated into the database as augmented training data. This iterative refinement or feedback loop training process involves continually updating the model with improved translations, thereby enhancing the overall quality of the translation engine.
\begin{CJK*}{UTF8}{bsmi} 
Our database contains three levels of training data:
\begin{enumerate}[left=0.2in]
    \item \textbf{Initial Training Data / Baseline Training Data:}\\
    Existing judgments translated by court translators.\footnote{Ideally, these translations should be examined, evaluated, and revised before being incorporated into the bilingual corpus as training data. Despite their high standard, they were not without issues, primarily because they were done without a coherent framework and clear guidelines. There is no mechanism for standardizing the translation of frequently used terms and expressions in judgments. For example, the basic term “judgment” was given two different translations as “判案書” and “判詞” instead of the official translation “判決書” in the Chinese translation of FACC1/2021 after two decades of experience translating judgments. Such inconsistency is only the tip of the iceberg, highlighting the need for considerable revision of existing translations before they can serve as reliable training data. However, we could not delay our project until all existing translations had been examined and revised; otherwise, it would never have commenced. As a result, we had to include existing translations in our corpus as initial training data.}
    \item \textbf{Post-Edited Data / Augmented Training Data:}\\
    These are revised translations of existing judgments.
    \item \textbf{Refined Training Data:}\\
    These are iterative refinements of revised judgments through incremental learning.
\end{enumerate}
\end{CJK*}

As the quality of training data has a direct bearing on the quality of output translations, evaluation plays a crucial role in producing ever-improving training data which will, in turn, produce ever-improving translations. For this purpose, we have developed a translation evaluation metric specifically for Hong Kong judicial judgments, named the \textbf{\textit{ACS score}}. We have conducted human evaluations of the TAP system's effectiveness. This metric includes three dimensions: 
\mbox{\textbf{A}ccuracy in Legal Meaning,} \mbox{\textbf{C}oherence and \textbf{C}ohesion in Textual Structure,} \mbox{\textbf{S}tyle Appropriateness.}

We have conducted a comprehensive analysis, providing clear insights into how different agents within the TAP system interact and communicate, and how they contribute to the overall system performance. This analysis aids subsequent research in making targeted improvements and innovations in this field, offering new perspectives for future studies.

\section{The CFA Judgement Corpus 97-22}

\subsection{Judgment document collection}

This study collected 344 Chinese judgments and their English versions from the Hong Kong court website between 1997 and 2022, constructing a corpus named \textit{CFA Judgement Corpus 97-22}, which has been open-sourced on Hugging Face\footnote{\url{https://huggingface.co/datasets/xxuan-nlp/CFA_Judgement_Corpus_97-22}}: 

\begin{figure}[htbp]
    \centering
    \includegraphics[width=0.95\textwidth]{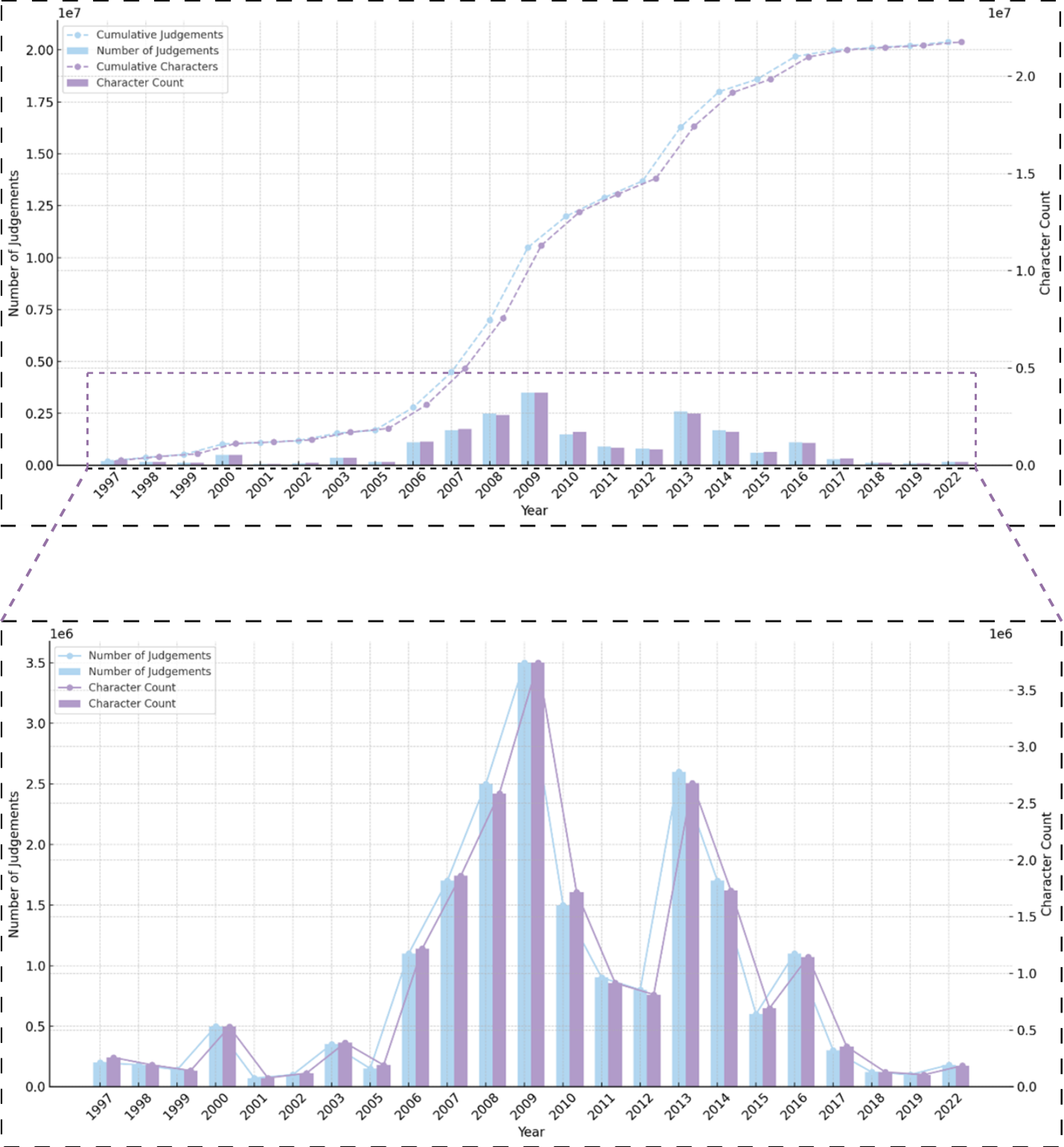}
    \caption{CFA Judgement Corpus 97-22 Data Statistics Chart.}
    \label{fig:trend_chart1}
\end{figure}

Figure~\ref{fig:trend_chart1} below shows the trend of the cumulative number of judgments in the \textit{CFA Judgement Corpus 97-22} over the years (top chart) and the annual distribution of judgments (bottom chart). 

The left vertical axis represents the number of judgments per year, while the right vertical axis indicates the corresponding character count for each year. The height of the bars represents the number of judgments each year, and the line connecting the peaks illustrates the trend in the character count of the judgments. The \textit{CFA Judgement Corpus 97-22} contains a total file size of 28,819.61~KB, encompassing 21,766,836 characters. 

This figure clearly reflects the yearly trends in the number of judgments and character counts. Since 2000, both the number of judgments and character counts have significantly increased, peaking between 2006 and 2013. This trend indicates a growing number of judgments issued by the Hong Kong Court of Final Appeal over time, correlating with increased case complexity and legal demands.

\subsection{Pre-processing}

\begin{figure}[htbp]
    \centering
    \includegraphics[width=0.95\textwidth]{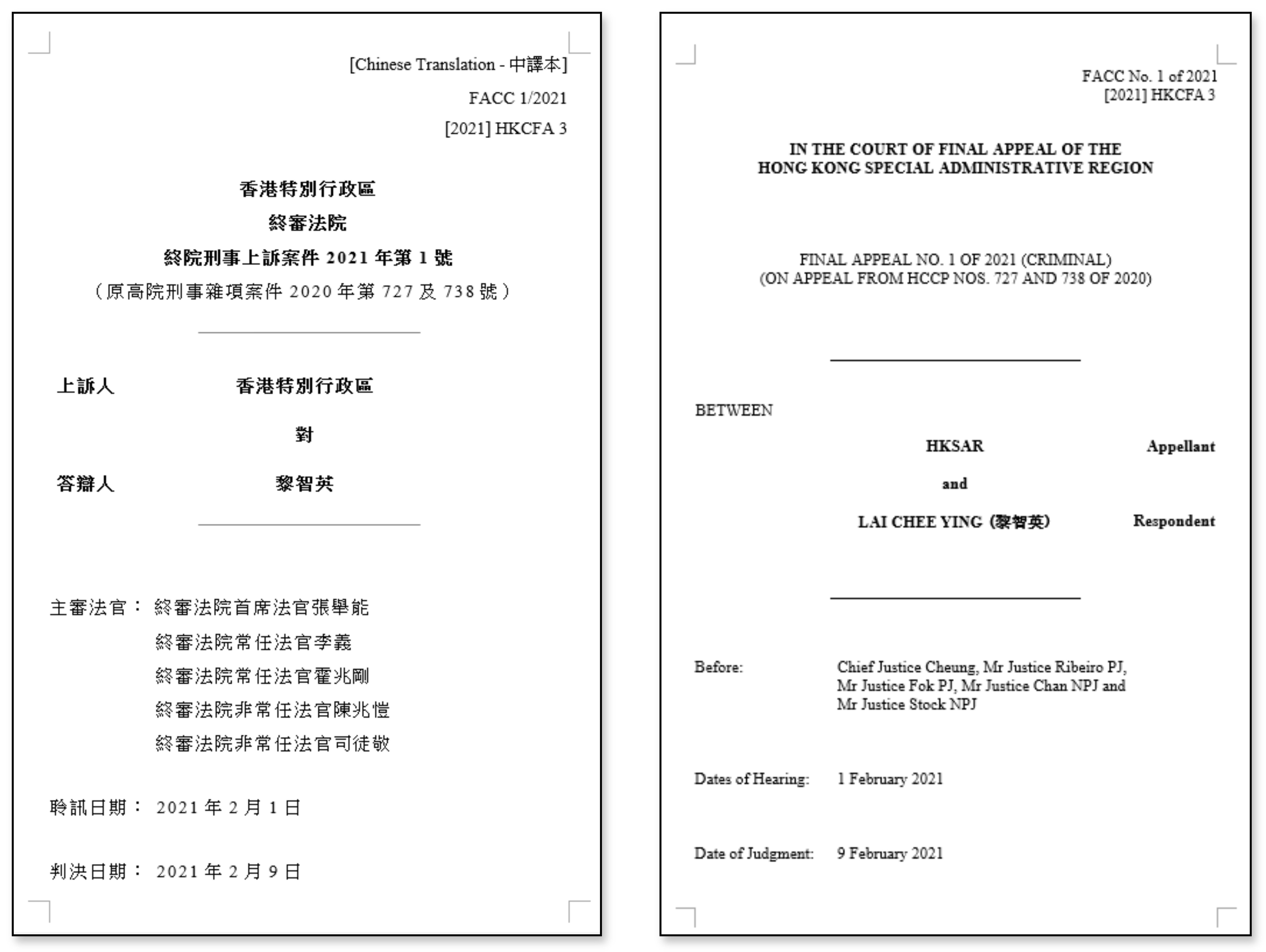}
    \caption{Judgment Format, using "HKSAR - Court of Final Appeal - Final Appeal Criminal Case No. 1 of 2021" as an example.}
    \label{fig:judgment_format}
\end{figure}

\begin{figure}[htbp]
    \centering
    \fbox{\includegraphics[width=1.0\textwidth]{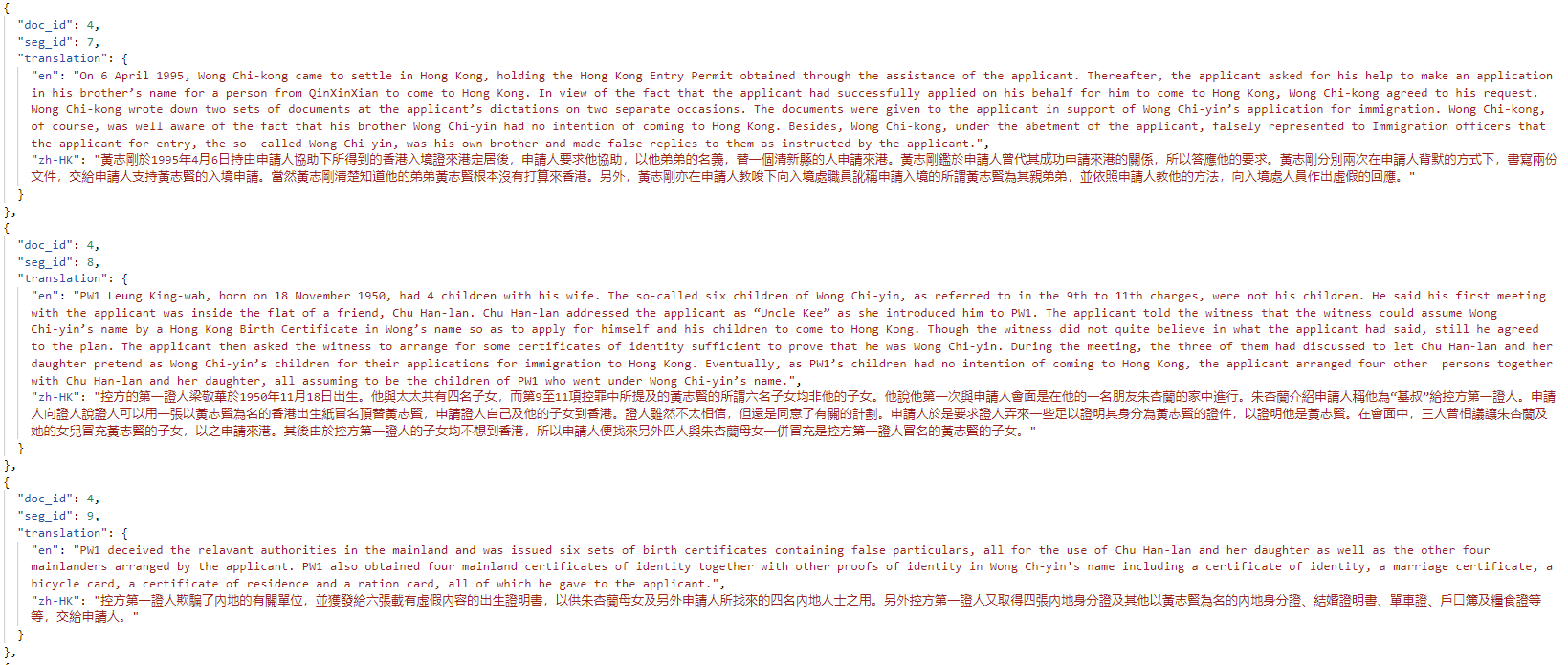}}
    \caption{Paragraph-Level English-Traditional Chinese Parallel Corpus.}
    \label{fig:parallel_corpora}
\end{figure}

In this section, we will describe the preprocessing process that constitutes the \textit{CFA Judgement Corpus 97-22}. This process includes reading, segmenting, and cleaning the \texttt{.docx} format judgment documents (as shown in Figure~\ref{fig:judgment_format}) to generate structured text in \texttt{Json} format suitable for analysis and machine learning tasks. This results in paragraph-level English-Traditional Chinese parallel corpora (as shown in Figure~\ref{fig:parallel_corpora}) stored in \texttt{Json} format (as shown in Figure~\ref{fig:json_format}).

To achieve paragraph segmentation and text alignment, we currently use Python combined with regular expressions for paragraph division and manual proofreading, including the removal of empty lines, spaces, and tab characters. The choice of regular expressions is due to their relative simplicity and quick application to small-scale corpus processing. However, to more efficiently handle large-scale corpora and improve the accuracy of segmentation and alignment, we plan to introduce machine learning-based sentence segmentation models in the future. This preprocessing step ensures that the generated data is clean and structured, thereby enhancing the accuracy and efficiency of subsequent analysis and machine learning models.

\subsection{Constructing the CFA Judgement Corpus 97-22}

In this subsection, we introduce the process of constructing the CFA Judgement Corpus 97-22. Using a preprocessing procedure, we generated paragraph-level bilingual pairs from .docx format judgment documents and created a paragraph-level English-Traditional Chinese parallel corpus stored in JSON format.

\begin{figure}[htbp]
    \centering
    \fbox{\includegraphics[width=0.75\textwidth]{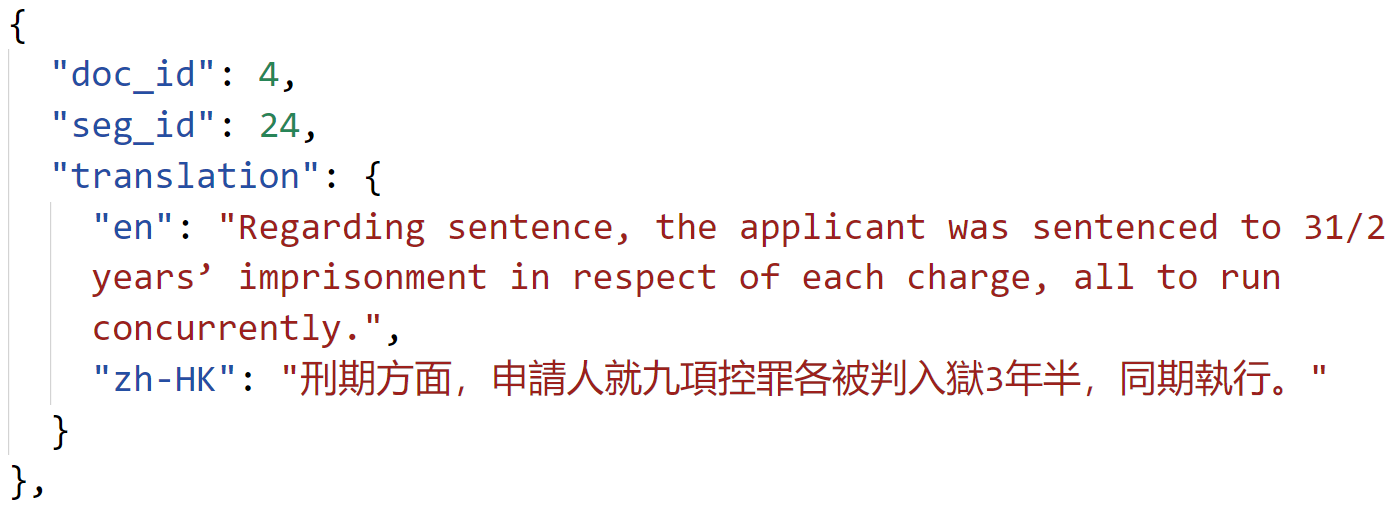}}
    \caption{A Single JSON Data Unit of Paragraph-Level English-Traditional Chinese Parallel Corpus.}
    \label{fig:json_format}
\end{figure}

The specific JSON format is shown in Figure~\ref{fig:json_format}. In this format: \texttt{"doc\_id"} represents the judgment document number; \texttt{"seg\_id"} represents the paragraph number; \texttt{"en"} represents the English text; and \texttt{"zh-HK"} represents the Traditional Chinese text (Hong Kong).

The corpus will be used as data sets for tuning and evaluating our MT and HMIT systems for judgments. Given the size of the corpus, we will initially focus on evaluation to ensure that the systems are performing effectively with the available data. However, due to the large volume of the corpus and the complexity of the tasks, there is a tremendous need to develop additional data to fully leverage the potential of our systems. This need for more comprehensive data sets underscores the importance of continuous data expansion and refinement to enhance the performance and accuracy of our MT and HMIT platforms.

\newpage 
\hmit{HMIT platform}

\setcounter{section}{0}

\section{Overall design}

The proposed project aims to develop an HMIT platform that utilizes a well-trained MT engine to facilitate the translation of judgments between English and Chinese. The methodology involves selecting cases for translation, building a bilingual parallel corpus as data for MT training/tuning, and using the best available MT engine for translation. The process includes feeding good data into the MT engine, post-editing its initial translations, and storing revised translations as new training data in the parallel corpus and exporting them to translation memory (example base) to further improve the MT. 

This cycle of human-machine interaction for post-editing repeats until the translations meet the required standards, ensuring required translation quality and continuous improvement of the MT engine, with the support of an external bilingual dictionary (defaulting to: Combined DOJ Glossaries of Legal Terms\footnote{\url{https://www.glossary.doj.gov.hk/}}).

The initial stage of our project involved preparing training data mainly from the bilingual judgments of criminal cases of the period of 1997--2021 available on the website of the Judiciary.\footnote{The initial bilingual text corpus contained only 214 pairs of English-Chinese judgments (1997--2021) which were increased to 333 as of June 2023: (1) 92 English criminal appeal judgments and their Chinese translations of the Court of Final Appeal; (2) 1 English criminal appeal judgment and its Chinese translation, and 236 Chinese criminal appeal judgments and their English translations of the Appeal Court of High Court; (3) 2 Chinese criminal trial judgments and their English translations of the Court of First Instance of High Court; and (4) 2 Chinese criminal trial judgments and their English translations of the District Court. As the initial core database was much too small to train the translation model to achieve the expected performance, two sub-datasets were subsequently built by web-mining from the following sources: (1) Parallel texts of about 2000 bilingual law texts from Hong Kong e-legislation (\url{https://www.eligislation.gov.hk}); and (2) Additional pairs of bilingual judgments from 763 judgments of the Judiciary (\url{https://www.judiciary.hk/en/home/index.html}). In addition to the bilingual judgments, a third data subset of 61,286 pairs of terms in English and Chinese was built.}

They were used to train the MT engine in use to underlie the HMIT platform, which allows for various post-editing operations to refine the MT outputs. The platform employed a novel non-monotonic alignment approach to handle cross-alignment in bilingual legal texts, enhancing the alignment accuracy of the bilingual corpus. The chosen MT engine, NiuTrans,\footnote{NiuTrans was developed by the Natural Language Processing Group at Northeastern University in China. Originally based on a statistical model, it has evolved to a deep neural network technology since 2016. At the time when we were choosing an automated translator for our platform, NiuTrans was by far the most powerful engine using Chinese as a pivot language. Consequently, it was the obvious choice for our automated translation engine.} was trained by its developer with multilingual big data including Chinese-English bitexts, supported by technical infrastructure for online access and post-editing, and is to be further fine-tuned with the aligned corpus for use in our HMIT platform.

This HMIT platform is intended to focus on user-oriented visualization for necessary bitext reading for post-editing and minimize text operations for the editing, leveraging both rudimentary online interactions through pop-ups and pull-downs and also advanced techniques like machine-learned lexical choices and word reordering. It acquires post-editing results as feedback to keep improving the MT engine's performance continuously.

On top of conducting post-editing, a user also plays the role of teaching the system to learn to translate better, and the system learns from its users' inputs to improve its translation, in a way of helping each other through interaction.

\begin{figure}[htbp]
    \centering
    \includegraphics[width=0.95\textwidth]{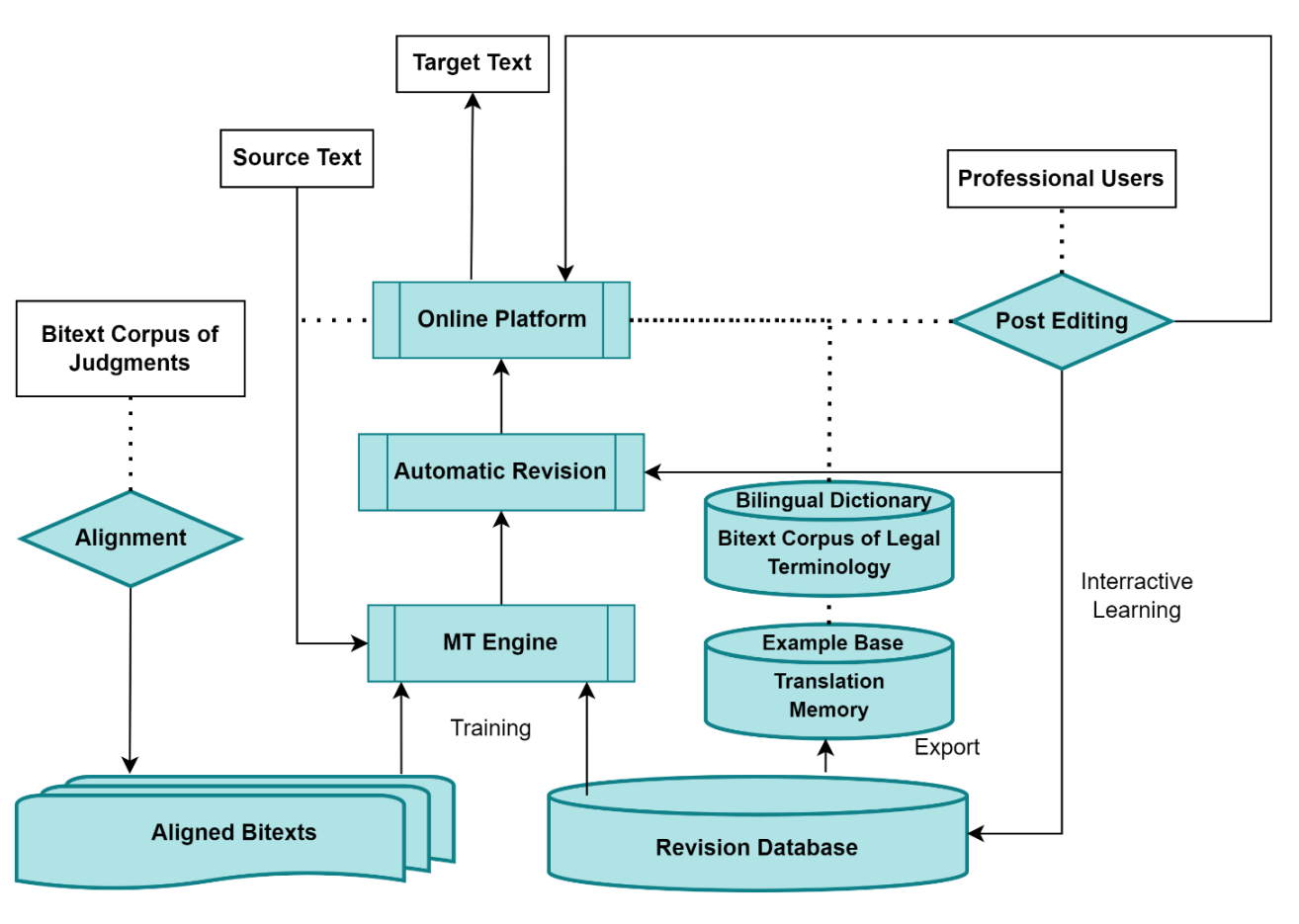}
    \caption{Workflow of platform operation.}
    \label{fig:workflow_diagram}
\end{figure}

We call this procedure interactive learning. The workflow diagram in Figure~\ref{fig:workflow_diagram} illustrates its components and how they are organized into one to support post-editing and interactive learning, highlighting the project's comprehensive approach to achieving high-quality legal translations. 

The project makes use of this platform to translate criminal appeal judgments of the Court of Final Appeal and the cited precedents, ensuring that the translations are accurate and efficient.

\section{Use of Large Language Models}

Before ChatGPT was first launched in November 2022, we were developing a bilingual translation system for proofreading using NiuTrans, which provided a comprehensive range of Application Programming Interfaces (APIs) with various operations for terminology and memory databases. These capabilities allowed us to build the prototype of the HMIT platform for demonstration purposes. 

As it turned out, however, the NiuTrans Engine did not lend us full support to enable a custom translation model as expected. In particular, its available services did not include a channel of deep learning using the post-edited translations for further improvement of its translation engine. Hence, we had to explore a novel approach by incorporating available well-trained large language models (LLMs). 

The rapid and drastic development of AI technology in recent years, especially the increasingly powerful versions of ChatGPT, brought about a great chance for more powerful and versatile translation technology and prompted us to realize the potential of LLMs in handling complex legal language and improving translation quality. As a result, in October 2023, we decided to transition our translation engine from NiuTrans’ neuro model to an LLM-based translation engine.

LLMs have revolutionized not only the field of natural language processing (NLP) but the entire Artificial Intelligence (AI). They are typically pre-trained on massive text data, so as to learn to predict the next word in a sentence \cite{brown2020language, chowdhery2022palm, fan2023bloom, team2023gemini, touvron2023llama,  bai2023qwen, anil2023palm2}. After pre-training, they are fine-tuned with instructions, through a process known as Supervised Fine Tuning (SFT) or Instruction Tuning (IT), so as to turn their capacity of language understanding into capability of following and executing human instructions \cite{sanh2021multitask, wei2021finetuned, tay2023chatgpt, longpre2023flan, shen2023alignment, chung2024large, wang2024multimodal}. 

Interestingly, recent research leveraging the exceptional abilities of LLMs shows that synthetic datasets generated by these models can also be used for this step ~\cite{luo2022biored, wang2023pandalm, wu2023llm, li2024llm, lyu2024retrieve, yue2024less}. Additionally, the performance of these models can be further enhanced by Reinforcement Learning from Human Feedback (RLHF), an approach to fine-tuning using feedback from humans or other large language models for rating the quality of model outputs \cite{ouyang2023llm, hejna2023contrastive, rafailov2024direct,  ethayarajh2024kto, hong2024feedback}. 

However, evaluating these large language models is a complicated task, which typically involves automated metrics and human judgment \cite{jiang2023identifying, hendrycks2023catastrophic, liang2024debatrix, wu2024self}. Other challenges to these models include efficient training \cite{hu2021lora, dettmers2024qlora}, fairness \cite{li2023fairness,cherepanova2024fairness}, and hallucinations \cite{zhang2023hallucinations, maleki2024ai}, for which there have been active areas in the frontier of ongoing AI research.

For this project, our switch to the application of Large Language Models (LLMs) has focused on utilizing the state-of-the-art LLMs as the core of our Machine Translation (MT) engine for translating Hong Kong judicial judgments. We have gained a better understanding of the advantages of LLMs. LLMs have a more advanced understanding of context due to their larger training datasets and complex architectures, maintaining context over longer passages and handling nuances better. They can adapt to a wider range of text styles and domains (e.g., legal, medical, casual conversations) because they are trained on diverse datasets and can also be fine-tuned for specific tasks or industries. Additionally, LLMs provide more consistent translations, maintaining high quality across various lengths and complexities of texts. They are better at resolving ambiguities and choosing the most appropriate translation based on broader context. LLMs excel in translating idiomatic expressions and colloquialisms, understanding cultural context better due to their extensive training data. Furthermore, they can be fine-tuned on specific datasets to improve performance in certain areas, making them highly customizable for specific needs. Finally, LLMs are more capable of integrating feedback and improving over time through iterative learning processes. 

These advantages, particularly continuous learning from feedback loops via post-editing, make our translation engine especially useful for translating complex texts of judicial judgments.

\section{Multi-agent system}

Another significant feature of our platform is that its Machine Translation (MT) engine has been transitioned from a single-agent system to a multi-agent system. Intelligent agents (IAs) are designed to understand their environment, make informed decisions, and take appropriate actions accordingly \cite{wooldridge1995intelligent}. The capabilities of Large Language Models (LLMs) align closely with these key characteristics, and their application can significantly advance the development of Multi-Agent Systems (MAS) for applications in various contexts. Compared to single-agent systems (SAS), MAS address complex problems or simulate intricate real-world environments through the collaboration of multiple IAs \cite{guo2024multiagents}. 

SAS based on LLMs have demonstrated exceptional cognitive abilities \cite{sumers2023cognitive}. Their development focuses on constructing internal mechanisms and enabling various interactions with the external environment. LLM-driven SAS demonstrate capabilities such as decision-making, which allows an agent to decompose a complex task into smaller sub-goals, tackle each one step-by-step, and learn from past experiences to improve decision-making skills \cite{khot2022decomposed, yao2024tree, shinn2024reflexion}. They can also utilize external tools and resources to accomplish various tasks effectively in diverse and dynamic environments \cite{li2023api, ruan2023tptu, gao2023retrieval}. Furthermore, they incorporate memory capabilities that encompass both short-term contextual learning and long-term external vector databases, enabling them to store and retrieve information over extended periods, maintain contextual coherence in text production, and enhance learning through interactions \cite{lewis2020retrieval, wang2023avalon, dong2024incontext}.

In contrast, MAS emphasize effective communication and interaction among agents with unique characteristics and the process of their collective decision-making. Multiple autonomous agents handle more dynamic and complex tasks through communication and collaboration with one another while maintaining their own unique strategies and behaviors \cite{guo2024multiagents}. Recent research has shown promising results of this approach in various fields such as software development \cite{hong2023metagpt}, multi-robot collaboration \cite{mandi2024roco}, scientific experiments \cite{du2023multiagentdebate}, and scientific debates \cite{xiong2023interconsistency}. Additionally, LLM-based multi-agent systems (LLM-MAS) play a crucial role in world simulation for social sciences, gaming, psychology, economics, and policymaking, (re)enacting various roles and perspectives through agents’ role-playing \cite{park2022socialsimulacra, park2023generativeagents, xu2023languageagents,li2023masquerade, mukobi2023welfarediplomacy, liang2023mathgpt} proposed a method of utilizing multi-agent debate in machine translation, although it was limited to the sentence level. In this research, we will focus on the paragraph level of legal judgment translation and proofreading, employing LLM MAS with translation/proofreading memory embeddings to address this complex and challenging task in machine translation, in hopes of enhancing the automation level of judgment translation in Hong Kong.

\section{Implementation of an MAS for Hong Kong Legal Judgment Translation}

\begin{figure}[htbp]
    \centering
    \includegraphics[width=1.0\textwidth]{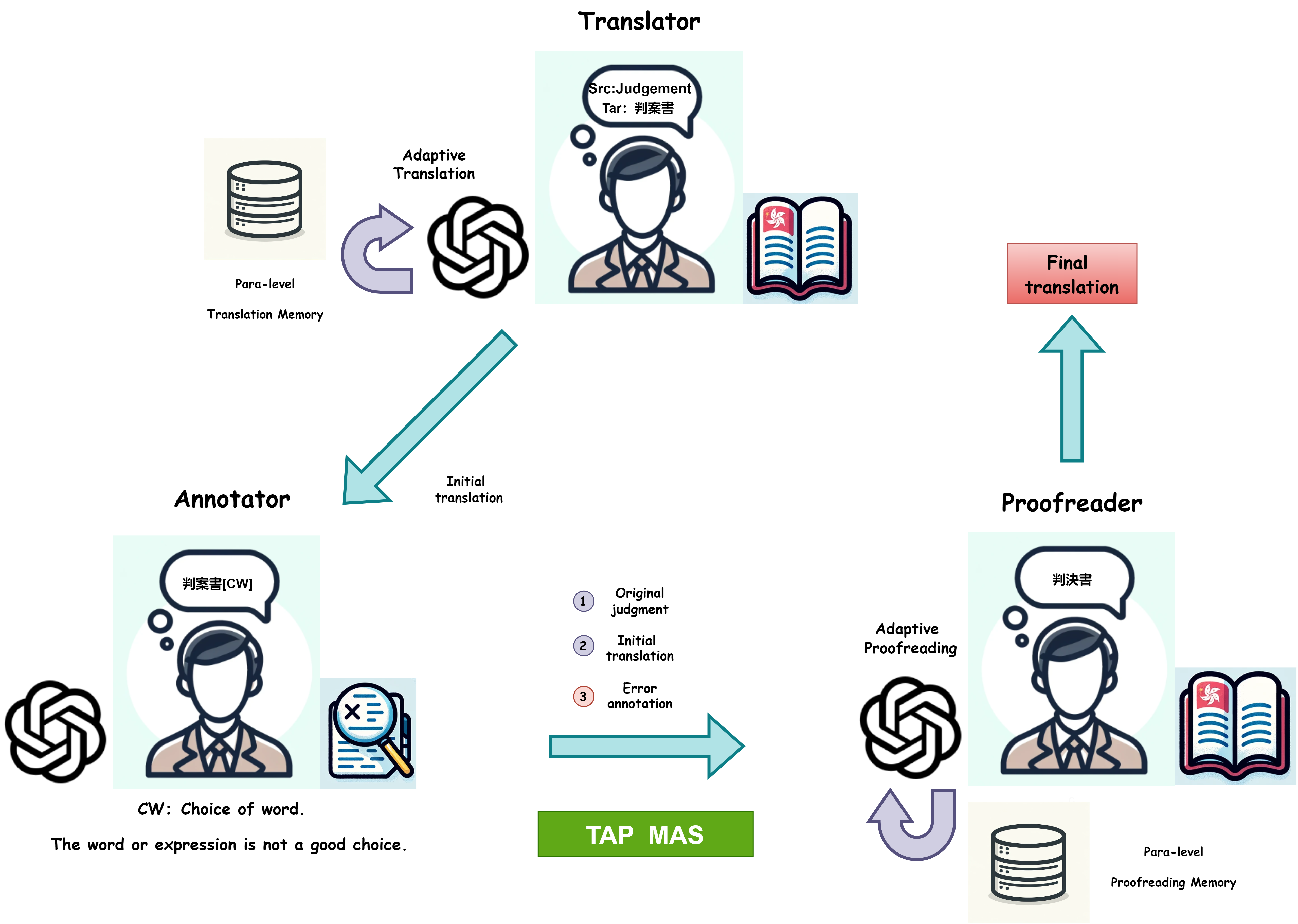}
    \caption{TAP MAS: a Multi-Agent System for Translating and Proofreading Hong Kong Legal Judgments.}
    \label{fig:tap_mas}
\end{figure}

We have established a virtual professional studio of MAS as the underlying MT engine to support the HMIT platform for Hong Kong legal judgment translation and proofreading. Its overall architecture is given in Figure~\ref{fig:tap_mas}. The roles of its three agents are Translator, Annotator, and Proofreader. Following these typical roles in translation, we call it TAP MAS, or simply TAP. 

This model simulates the entire translation process of a judgment (or any text), with these agents in different roles co-working together to ensure the quality and consistency of the final product throughout the whole translation process. In the following subsections, we will present the roles and core collaboration strategies of its agents, and its workflow to carry out translation tasks.

\subsection{Roles of Agents}

To simulate the entire translation process of a judgment, the three agents in TAP take various roles as follows, according to each one’s responsibilities.

\begin{enumerate}
    \item \textbf{Translator:} Responsible for accurate translation of the judgment from English to traditional Chinese, ensuring the preservation of its legal meanings, terminology, and the tone of the judgment, ensuring the accuracy and completeness of the translation according to the context and background, and also ensure consistency in legal terminology throughout the translation process so as to avoid confusion or misunderstanding.

    \item \textbf{Annotator:} Responsible for marking errors in the Translator’s translation according to the multi-level translation evaluation annotation standard (Proofread Codes, see Appendix Table 1). The errors to be annotated include but are not limited to the following types: (1) Accuracy errors: Incorrect word choice, failure to preserve information structure, meaning alteration, mistranslation, different meaning, unclear meaning, omissions, over-translation, overly literal translation, and insufficient translation; (2) Grammatical errors: Misuse of article, determiner, modal verb, numeral, preposition, and even punctuation; wrong part of speech, misspelling or mistaken character, ill-formed structure, subject-verb disagreement, tense error, and wrong word order; (3) Usage and style errors: Wrong collocation, inappropriate meaning, incorrect connective, inconsistent usage, unnatural expression, reference problems, redundancy, inappropriate style, and problematic transition. The Annotator’s role is to provide detailed error annotations and modification suggestions to the Proofreader.

    \item \textbf{Proofreader:} Responsible for correcting and revising the initial translation from the Translator according to the Annotator's error annotations, conducting the final review, and finalizing the translation.
\end{enumerate}

Through the collaborative work of the three agents in these roles, TAP seeks to maximize the accuracy, completeness, and professionalism of judgment translation up to a quality level to meet the rigorous requirements of the legal field. To examine the realism of TAP’s translation process simulation, we use GPT-3.5 Turbo as the agent LLM for all three roles. 

To ensure that the LLM fully understands the task content, avoids hallucinations, and produces precise and concise outputs, we have carefully formulated respective role prompts for these roles, as presented in Figure~\ref{fig:tap_mas_prompt}. Technically, we have detailed 30 subcategories of translation error in the prompts for the Annotator and Proofreader, corresponding to the multi-level translation evaluation annotation standard (Proofread Codes) developed by Hong Kong judgment translation experts. 

The experiments we carried out to test TAP verified that this approach to error annotation feedback can guide the LLM effectively in correcting mistakes in translation, supporting and inspiring future research in this field.

\subsection{Core Strategies of Agent Collaboration}

Agent capability acquisition is a key process in LLM-MAS that enables agents to learn and evolve incrementally in a dynamic manner. In TAP, this acquisition process is crucial, ensuring the agents continuously enhance their ability and performance. 

Two fundamental issues need to be handled in this process: one is how the agents receive feedback of various types, and the other is how they adjust themselves accordingly in order to carry out their roles to address complex problems.

\subsubsection{Feedback Types}

There are two basic types of feedback in TAP, as follows. (1) Feedback between agents for collaborative interaction: An agent receives feedback from another as response to or as judgment about its output through communication between agents. This form of feedback promotes cooperation and information sharing among agents, for the purpose of optimizing the overall performance of the whole MAS. (2) Human feedback: This kind of feedback from humans serves the purpose of ensuring what the LLM-MAS in question does or produces aligns well with human knowledge (such as the expertise of experts) and/or preference (such as translation style). This feedback mechanism aims at helping the system understand and meet user needs properly, in hopes of enhancing the accuracy and naturalness of translation.

\subsubsection{Self-Adaptation}

To further enhance translation and proofreading performance, TAP incorporates two self-adaptation strategies. One is a memory module that allows the agents to store and retrieve their interaction records in the past, including translation and proofreading memory, and feedback information. It enables a continuous learning mechanism that allows the agents to improve their performance by utilizing available historical data. 

The other is self-evolution, which allows the agents to adjust how they perform their roles via learning from their interactions with humans using feedback or communication logs. This strategy may lead to continuous changes in working methods and subtasks to fulfill the roles of the agents, aiming at further improvement of the overall intelligence and efficiency of the MAS.

\subsection{TAP MAS Workflow}

By virtue of the above strategies, TAP aims at efficient and accurate translation of complex legal documents via a highly collaborative process. This strategy relies on the close cooperation of its three agents, which play specific roles as specified in respective prompts. This multi-layered collaborative approach ensures meticulous handling at each stage and is thereby expected to have a high potential for enhancing the overall translation quality at the system level. The workflow for judgment translation and proofreading in the TAP MAS includes two main phases, namely preparation and execution, to be detailed below.

\subsubsection{Preparation}

The system first allows its users to make their choices about what to use, such as selecting the agents for the three roles from available ones. Currently, there are four choices for the Translator agent, namely, NiuTrans, GPT-3.5-turbo, GPT-4-turbo, and GPT-4o. The first choice can only be used as an MT system, while the latter three are also the available LLM choices for the Annotator and Proofreader. We will continue to integrate more upcoming LLMs into this system as available choices. In addition, a user also needs to decide the translation direction by selecting the source and target language, for which the default choice is from English to Traditional Chinese, and choose the terminology database, for which the default choice is the Combined DoJ Glossaries of Legal Terms (but another choice is provided for users to import their own), in addition to the translation and proofreading memory database.

\subsubsection{Execution}

\begin{figure}[htbp]
    \centering
    \includegraphics[width=1.00\textwidth]{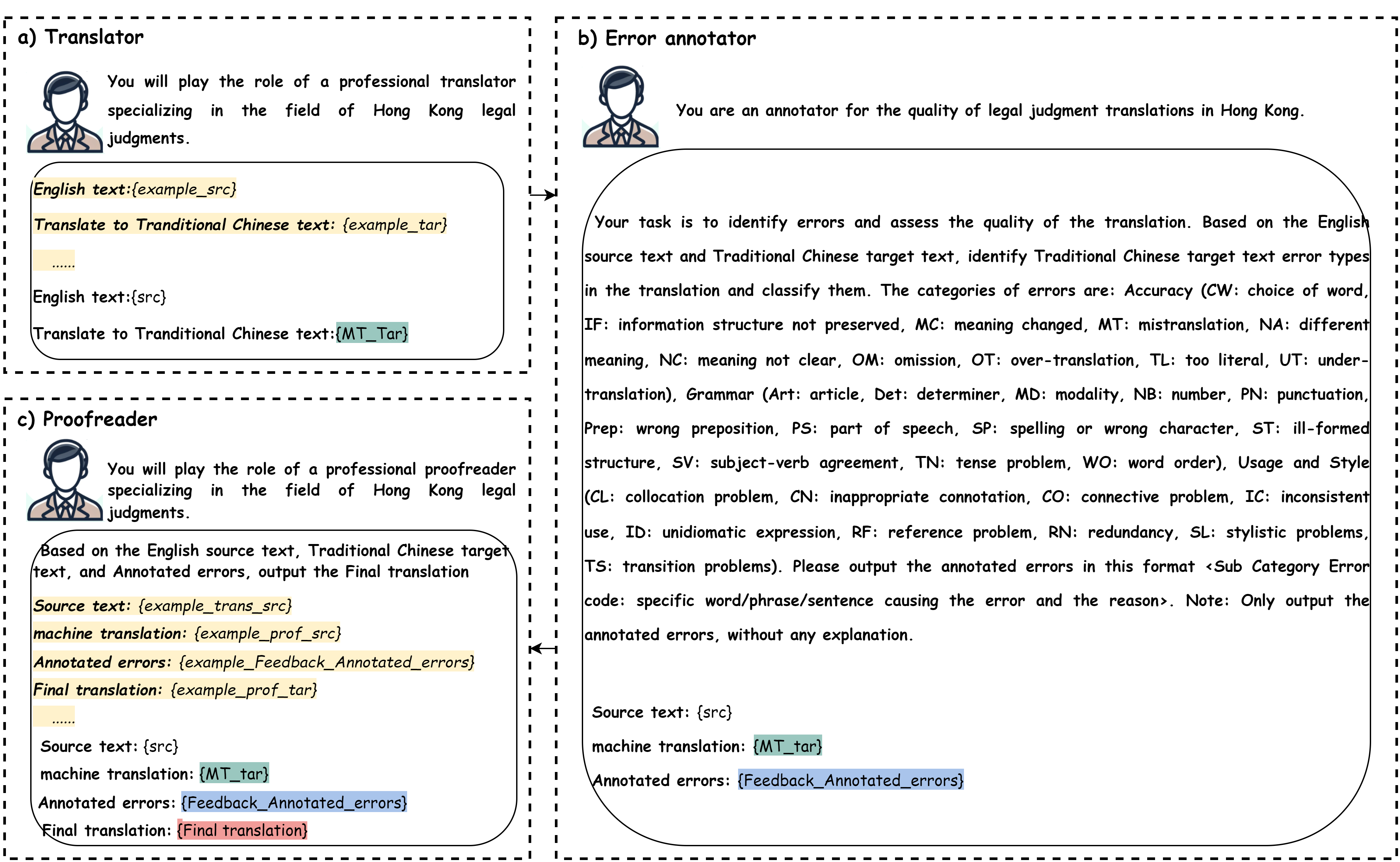}
    \caption{Illustration of few-shot prompts used in TAP MAS (Green/blue/red highlights indicate the outputs of the T/A/P Agent, respectively).}
    \label{fig:tap_mas_prompt}
\end{figure}

The process of TAP MAS execution subsumes three distinct sub-stages according to the roles of the three agents, namely, translation, error annotation, and proofreading, as presented in Figure~\ref{fig:tap_mas}. Figure~\ref{fig:tap_mas_prompt} illustrates the details of the prompts for the three agents. Note that using prompts has been of the most vital importance in almost all applications of LLMs. Recent studies demonstrate that an ideal prompt can significantly boost the model performance and unlock surprising model capabilities \cite{kojima2022zeroshot, ge2022extensible}. 

Designing effective prompts to guide TAP agents in understanding their roles, task requirements, and specific operational steps is crucial in improving the efficiency and output quality of the entire system. As illustrated in Figure~\ref{fig:tap_mas_prompt}, the prompts for different TAP agents comprise both role prompts and task prompts. To further optimize TAP’s performance, the two agents, Translator and Proofreader, also incorporate example prompts (few-shot; underlined in Figure~\ref{fig:tap_mas_prompt}). We have identified the optimal setting of $K=5$, which coincides with that in \cite{zhang2023prompting}. The complete list of prompts used in TAP MAS is provided with all details in Tables~\ref{tab:AT2}--\ref{tab:AT6} in the Appendix.

Putting these Figure~\ref{fig:tap_mas} and Figure~\ref{fig:tap_mas_prompt} together, we can see how these agents work, from receiving an original judgment (mostly in English) as input to outputting a final translated judgment (in Traditional Chinese). More technical details of this process are as follows in three phases. In the first phase, the translation process begins by dividing a document-level judgment input into multiple paragraphs, each of which then undergoes preliminary translation by the Translator agent. This agent uses GPT-3.5 Turbo as its agent LLM with a few-shot prompting strategy to feed it a certain amount of translation memory for tuning purposes. To ensure consistency and accuracy, the few-shot prompts are generated using the Physical Neighbor Sampling (PNS) method proposed in this paper, which extracts the five pairs of source and target paragraphs from the translation memory database that are geographically closest to the current paragraph under translation. These pairs serve as contextual examples for the LLM to learn from, i.e., to perform fine-tuning using such examples in hopes of translating the current paragraph better than without such examples for tuning. Once completed, this preliminary translation is sent to the next phase for translation error annotation, during which the Annotator agent reviews each translated paragraph, identifies and annotates translation errors in terms of the translation error types stipulated in the Proofread Codes, and assigns corresponding error codes accordingly. These annotation outputs are stored with respective source and target as triplets in the proofreading memory database for later use. Finally, in the next phase of proofreading, the Proofreader agent revises the annotated preliminary translations output from the Annotator. This agent is an LLM leveraging proofreading memory via few-shot prompting that selects the five closest triplets from the proofreading memory database using the PNS method. It corrects the preliminary translations in terms of the errors identified by the Annotator and consequently produces the final translation. This result is then added to the proofreading memory database for use later in few-shot prompts to help translate other incoming source paragraphs, in order to maximally ensure high quality of translation and consistency in style throughout the entire document.

\section{Evaluation}

As discussed above, an MT engine needs to be employed to support our HMIT platform that aims not only at supporting human post-editing, which is still indispensable to further enhance the initial translation by MT up to a required quality level, but also at minimizing human effort spent on post-editing. It is straightforward to see that the higher the quality of the initial translation, the less human effort is needed for post-editing. The translation quality of the MT engine in use is one of the most crucial factors that determines the usefulness of the HMIT platform. In this section, we report both automated and human evaluation of our MT engine.

\subsection{Automated evaluation}

Automated evaluation of an MT system is conducted by applying available authentic automated metrics to compute quality scores for its translation output by contrasting the output with the gold standard answers in a given bilingual text dataset.

\subsubsection{Metrics}
    The metrics that we adopted for our evaluation are the following three that have been the most popular in recent years for automated MT evaluation: 
    (1) xCOMET-XL, a version of xCOMET, which is a state-of-the-art learned metric for various levels of evaluation \cite{guerreiro2024xcomet}; 
    (2) Reference-free quality evaluation model COMETKiwi, developed as a reference-free evaluation metric \cite{rei2022cometkiwi}; 
    (3) Unified MT quality evaluation model wmt22-unite-da, a unified MT quality evaluation model \cite{guttmann2024comet}.

\newcounter{partnumber} 
\setcounter{partnumber}{1} 

\subsubsection{Experiments for Evaluation}

\begin{enumerate}[label=(\arabic*), leftmargin=2em] 
    \setcounter{enumi}{0} 
    \item \textbf{Test Set} \\
    We selected the bilingual texts of the judgment “HKSAR - Court of Final Appeal - Final Appeal Criminal Case No. 1 of 2021” from the CFA Judgement Corpus 97-22 dataset as our test data. We may refer to this case as FACC 1/2021 henceforth for brevity. The main reasons to choose it include its availability and our team members’ familiarity with it. After the preprocessing as described in Part \Roman{partnumber}-section 7.2, including paragraph-level segmentation and manual alignment, the whole test set consists of 200 paragraph-level source-target pairs. According to the \href{https://platform.openai.com/tokenizer}{Tokenizer}\footnote{\url{https://platform.openai.com/tokenizer}\label{fn:tokenizer}} we used for the statistics, its source text consists of 12,029 tokens (57,926 characters) in English.
\end{enumerate}

\begin{enumerate}[label=(\arabic*), leftmargin=2em] 
    \setcounter{enumi}{1} 
    \item \textbf{Models} \\
    The LLM we used for all three agents in TAP MAS is GPT-3.5 Turbo. Our HMIT platform integrates two types of MT engines: NMT- and LLM-based ones. The former includes NiuTrans,\footnote{NiuTrans: Developed by the NLP Lab at Northeastern University, China, this neural network based model supports multiple language pairs, offering quality translation with nice efficiency and a flexible architecture. It is widely used in academia and industry.} Google Translate,\footnote{Google Translate: An online translation service developed by Google, based on powerful neural machine translation (NMT) technology. It covers over 100 languages and is continually updated to improve translation quality and fluency.} and DeepL,\footnote{DeepL: Developed by DeepL GmbH, it is renowned for its excellent translation quality and smooth output. Using advanced deep learning algorithms, it is particularly adept at handling complex sentence structures and semantic nuances.} and the latter GPT-3.5 Turbo,\footnote{GPT-3.5 Turbo \cite{openai_gpt35_turbo}: As one of the advanced LLMs developed by OpenAI in the ChatGPT series, it excels in tasks such as translation, conversation, Q\&A, and content generation. It reportedly outperforms GPT-3 and OPT-175B in multiple natural language processing tasks \cite{zeng2022glm}. Using GPT-3.5 Turbo grants a nice balance of economical efficiency and high quality of translation.} GPT-4.0 Turbo,\footnote{GPT-4.0 Turbo \cite{openai_gpt4_turbo}: Building on GPT-3.5, GPT-4.0 Turbo has undergone further optimization, significantly enhancing processing speed and response quality. It has stronger reasoning and comprehension capabilities, making it suitable for more complex language tasks and application scenarios.} and GPT-4o Turbo.\footnote{GPT-4o Turbo \cite{openai_gpt4o}: As the latest iteration, GPT-4o Turbo reaches new heights in accuracy and efficiency, further reducing response times and enhancing text generation and analysis capabilities, making it more suitable for high-demand professional applications.} In the future, we will continue to update and integrate state-of-the-art LLMs for users’ choices in our HMIT Platform.
\end{enumerate}

\begin{enumerate}[label=(\arabic*), leftmargin=2em] 
    \setcounter{enumi}{2} 
    \item \textbf{LLM Response Parameter Settings} \\
    Specifically, TAP is composed of the three agents whose LLM response parameter
    setting is the same as follows: \texttt{Temperature = 0}, \texttt{max\_tokens = 4,096}, \texttt{frequency\_penalty = 0} 
    and \texttt{presence\_penalty = 0}. However, the same parameter setting for the agents of
    different roles serves different purposes.
    
    \begin{enumerate}[label=\arabic*.] 
        \item \textbf{Translator:} A low \texttt{temperature} value like \texttt{0} indicates the use of a greedy decoding
        strategy in GPT-3.5 Turbo, making it as focused and deterministic as it can. Note
        that in greedy decoding, the model selects the token with the highest probability at
        each step of text generation. This setting limits the creativity and coherence of the
        agent, ensuring its determinism and consistency. The \texttt{max\_tokens} value \texttt{4096} is
        assumed to be long enough to handle paragraph-level legal judgment texts. The two
        penalty values are set to \texttt{0} to make the agent inclined to generate a text that tends
        to contain high-frequency words and look like the input, respectively.
        
        \item \textbf{Annotator:} The \texttt{temperature = 0} is to ensure the output rigor while the \texttt{max\_tokens = 4096} 
        is supposed to be sufficient for paragraph-level error annotation. The two
        penalty parameters are set to \texttt{0} in order to minimize repetitive error annotations and
        to make the error identification as accurate as possible.
        
        \item \textbf{Proofreader:} Similarly the \texttt{temperature = 0} is to ensure the rigor of the legal text
        proofreading output. The \texttt{max\_tokens} is set to \texttt{4096} to handle paragraph-level long
        texts. The two penalty parameters are set to \texttt{0} in order to avoid redundant
        corrections and to ensure comprehensive proofreading, respectively.
    \end{enumerate}

    These parameter settings are intended to tailor GPT-3.5 Turbo, as the agent LLM,
    to best fit the scenario of Hong Kong legal judgment translation and in particular, facilitate
    the expected effective and efficient collaboration between the three agents to cope with the
    complexity of judgment documents, so as to ensure the highest possible quality of the
    outcome of machine translation and automated proofreading in hopes of minimizing human
    efforts in a later phase of human proofreading using the HMIT platform.
\end{enumerate}

\subsubsection{Evaluation Results and Analysis}

The performance of TAP MAS with different role configurations, in terms of K-shot
example prompt (if applicable), for its three agents is reported in Table~\ref{fig:results_as_table}, according to our
evaluation using the bilingual texts of FACC 1/2021 as the test set and \texttt{XCOMET-XL} and \texttt{Wmt22-unite-da} 
as evaluation metrics. The configurations can be grouped into two categories for the purpose of
comparison, i.e., \texttt{MAS 1-5} as one and \texttt{MAS 6-10} as another. In each group, there is a baseline
(i.e., the one with the smallest number) and other variations on top of it for possible enhancement. In addition, 
manual error annotation by an expert in legal translation\footnote{The first author.} is also brought in to replace
the Annotator agent for comparison.


\begin{table}[h!]
\centering
\caption{Performance of TAP MAS with different configurations for the three agents (T: Translator, A: Annotator, P: Proofreader; X: not used)}
\label{fig:results_as_table}
\begin{tabular}{cccccc}
\toprule
\multirow{2}{*}{MAS} & \multicolumn{3}{c}{Agent: 0- vs 5-shot} & \multicolumn{2}{c}{Metric} \\
\cmidrule(lr){2-4} \cmidrule(lr){5-6}
 & T & A & P & XCOMET-XL & wmt22-unite-da \\
\midrule
1 & 0 & X & X & 0.2192 & 0.6172 \\
2 & 0 & X & 0 & 0.7635 (+0.5443) & 0.8574 (+0.2402) \\
3 & 0 & X & 5 & 0.8028 (+0.5836) & 0.8662 (+0.2490) \\
4 & 0 & LLM & 0 & 0.8466 (+0.6274) & 0.8664 (+0.2492) \\
5 & 0 & LLM & 5 & 0.8633 (+0.6441) & 0.8726 (+0.2554) \\
\midrule
6 & 5 & X & X & 0.8381 & 0.8745 \\
7 & 5 & X & 0 & 0.8330 (-0.0051) & 0.8709 (-0.0036) \\
8 & 5 & X & 5 & 0.8486 (+0.0105) & \textbf{0.8749 (+0.0004)} \\
9 & 5 & LLM & 0 & 0.8435 (+0.0054) & 0.8637 (-0.0108) \\
10 & 5 & LLM & 5 & \textbf{0.8669 (+0.0288)} & 0.8732 (-0.0013) \\
\midrule
11 & 5 & Manual & 0 & 0.8290 & 0.8662 \\
\bottomrule
\end{tabular}
\end{table}

\begin{enumerate}[label=(\arabic*), leftmargin=2em] 
    \setcounter{enumi}{0} 
    \item \textbf{Baseline Systems (MAS 1 and 6)} \\
    In these two baseline systems, only the translation module (namely, the Translator agent)
    is used, one with zero-shot and the other with 5-shot example prompt. The performance of the
    former is extremely poor in terms of its \texttt{XCOMET-XL} and \texttt{wmt22-unite-da} scores \texttt{0.2192} and
    \texttt{0.6172}, respectively, indicating that a single zero-shot translation module alone cannot provide
    acceptable translation quality and coherence when handling complex legal texts. Using 5-shot
    instead, however, MAS 6 shows a significant improvement in translation performance, pushing
    the two evaluation scores up by 3.8 and 1.4 times, respectively, and confirming the effectiveness
    of the LLM tuning strategy by means of using translation memory in the prompt for this particular
    Translator agent that plays the key role of translating legal judgments.
\end{enumerate}

\begin{enumerate}[label=(\arabic*), leftmargin=2em] 
    \setcounter{enumi}{1} 
    \item \textbf{Effect of Proofreader} \\
    When the Translator is of zero-shot (MAS 1), adding a Proofreader of either zero- or 5-shot
    (as in MAS 2 \& 3) significantly improves the system’s performance, and the latter improves
    more. When the Translator is of 5-shot (MAS 6), adding a Proofreader of zero- or 5-shot (as in
    MAS 7 \& 8) does not lead to any significant change in translation performance, despite the former
    decreases the scores while the latter increases them, both marginally.
\end{enumerate}

\begin{enumerate}[label=(\arabic*), leftmargin=2em] 
    \setcounter{enumi}{2} 
    \item \textbf{Effect of Annotator} \\
    In the four pairs of contrastive configurations (namely \texttt{MAS 2 vs 4}, \texttt{3 vs 5}, \texttt{7 vs 9}, and \texttt{8 vs 10}) 
    to compare using vs not using an Annotator agent in similar configurations, only two (among
    eight) pairs of evaluation scores (namely \texttt{wmt22-unite-da} scores for the last two pairs) show a
    negligible negative effect of the Annotator, while the rest show a positive improvement to various
    extents. As a whole, the proposed independent error annotation module appears to be effective in
    enhancing the overall performance of translating legal judgments. Notably, the Annotator is more
    effective for lower configurations (e.g., zero-shot T \& P), as shown by an improvement of
    \texttt{XCOMET-XL} scores by 10.88\% from \texttt{MAS 2 to 4}, but marginally effective for higher
    configurations with 5-shot T \& P.
\end{enumerate}

\begin{enumerate}[label=(\arabic*), leftmargin=2em] 
    \setcounter{enumi}{3} 
    \item \textbf{Effect of Translation Memory} \\
    In the five pairs of contrastive configurations (namely \texttt{MAS 1-5 vs 6-10}) to examine the
    difference of not using vs using translation memory, only one (namely \texttt{MAS 4 vs 9}) shows
    a negligible negative effect of adding translation memory via 5-shot example prompt, and all the
    others show a positive improvement in the overall translation performance.
\end{enumerate}

\begin{enumerate}[label=(\arabic*), leftmargin=2em] 
    \setcounter{enumi}{4} 
    \item \textbf{Effect of Proofreading Memory} \\
    In the four pairs of contrastive configurations (namely \texttt{MAS 2 vs 3}, \texttt{4 vs 5}, \texttt{7 vs 8}, and \texttt{9 vs 10}) 
    to examine the effectiveness of not using vs using proofreading memory (i.e., zero- vs 5-shot P), as a whole they confirm 
    an overall improvement of the system performance by 5-shot P. However, the two evaluation metrics show some inconsistency 
    in scoring for this improvement, e.g., \texttt{MAS 10} underperforms the baseline \texttt{MAS 6} in terms of their 
    \texttt{wmt22-unite-da} scores, despite a negligible difference between them. However, this tiny gap of scoring invites us 
    to rethink about the possible limit of automated metrics in reflecting the quality of translated legal judgments, given the 
    lack of semantic understanding in such metrics to handle the complexity of legal language, its accuracy in terms of semantics 
    and contextual applicability, and also the prominence of legal terminology. Legal judgments contain many specialized terms 
    and expressions that automated metrics usually treat as normal words and do not recognize their special values in translation. 
    Moreover, automated metrics are typically paired with the multidimensional quality metrics (\texttt{MQM}) framework 
    \cite{mariana2014mqm} for translation evaluation, but it remains a problem if this framework has incorporated all necessary 
    dimensions for assessing the quality of Hong Kong legal judgment translation, including accuracy in legal meaning, 
    appropriateness in style, coherence and cohesion in discourse structure, to name but a few. As many of them go beyond the 
    scope of existing automated metrics for MT evaluation and MQM-based human evaluations, we hence need a manual evaluation 
    by domain expert covering these three dimensions, which will be reported in Section~6.

    From the experimental results presented in Table~1, we can see that \texttt{MAS 10}, of the highest configuration 
    (i.e., 5-shot for both T and P in addition to A), gives the best overall performance with an \texttt{XCOMET-XL} score 
    higher than all others and a \texttt{wmt22-unite-da} score of insignificant difference from the highest, validating the 
    effectiveness of the three agents’ collaboration on translating Hong Kong legal judgments. Interestingly, its performance 
    even surpasses that of \texttt{MAS 11}. Comparing the performance of \texttt{MAS 4} and \texttt{11}, we are interested in 
    knowing if the Annotator agent could have done a better job than human annotation. Certainly, a lot of research in this 
    direction is yet to be conducted with much genuine efforts. To further verify the above experimental results, however, we at 
    least need a human evaluation on top of the automated evaluation as reported above.
\end{enumerate}

\section{Human Evaluation}
For human evaluation, we first need to formulate a scoring scheme for use to integrate a human
evaluator’s scores in various evaluation dimensions into one. The one we have developed
specifically for the translation of Hong Kong legal judgments is the legal ACS metric (or simply
ACS for brevity), whose formulation will presented in the next subsection, followed by the settings
and results of our human evaluation.

\subsection{Evaluation Metrics}
Aimed at a comprehensive, adequate and reliable evaluation of the translation quality of
Hong Kong legal judgments, the ACS metric is formulated as follows,
\begin{equation}
    I = \alpha A + \beta C + \gamma S
\end{equation}
where $A$, $C$, and $S$ are the scores in the three key dimensions of evaluation by a human expert
evaluator, namely, accuracy of legal meaning, coherence and cohesion in structure, and
appropriateness in style, and $\alpha$, $\beta$, and $\gamma$ are their respective weight coefficients 
according to the relative importance of these dimensions. Based on the experience and recommendation of domain
experts, these weights are set as follows for our manual evaluation of legal judgment translation:\(\alpha = 0.6, \beta = 0.3, \gamma = 0.1\). This setting recognizes the most fundamental role of the accuracy
of legal meaning as the key criterion in determining the quality of legal translation, i.e., how well
the legal meaning is accurately conveyed through translation, and accordingly gives it the highest
priority. The next most critical feature that differentiates the genre of legal text from all others is
the coherence and cohesion of the original legal text (30\%) and hence the preservation of
coherence and cohesion in legal translation is given the next highest weight, estimated half as that
of legal meaning. Slightly less influential than these two criteria but still significantly important to
translation quality is the stylistic appropriateness in the translation for maintaining the appropriate
legal style. The weight we considered appropriate for it is one third of that for coherence and
cohesion. 

Note, however, that these weights we adopted for this evaluation were an \textit{ad hoc} setting
empirically chosen according to our experience, and how to develop an optimal setting for legal
translation on top of this, of course, has to be reserved for future research.

\subsection{Setup}

Because of the high cost of human evaluation, it is unrealistic to use the whole FACC
1/2021 test set for this purpose due to our limited manpower and resources. Instead, we randomly
selected 10 segment-level pairs from it for three representative MT systems, namely GPT-4o as
baseline for comparison, our MAS 10 (of the highest configuration) and MAS 11 (involving manual
error annotation), to translate and then compare their performance. Given that the longest segment
is of 234 words (290 tokens/1,432 characters) in English and 414 words (580 tokens/460 characters)
in Traditional Chinese, and it is difficult for evaluators to follow the given criteria strictly when
assessing lengthy text pairs, we manually split the translated segments (in Traditional Chinese) into
sentences, resulting in 25 sentence-level pairs whose longest sentences are reduced to 91
words (96 tokens/486 characters) in English and 92 words (187 tokens/135 characters) in
Traditional Chinese (The token/character counts are based on statistics from the OpenAI
Tokenizer\footref{fn:tokenizer} used for this analysis.), before distributing them anonymously to human evaluators in
an evaluation table with segment number, sentence number, source text (English), and MT system
ID. The evaluators assessed the translations along the three quality dimensions according to their
experience of many years in legal translation and gave each translated sentence a subjective quality
score in the range of 0 to 10, with 10 for the highest translation quality and 0 for the lowest.

\subsection{Results and Analysis}

The human evaluation results are presented in Table~2, from which the most remarkable
observation we have is that both MAS 10 and 11 using GPT-3.5 Turbo as the LLM for the three
agents outperform GPT-4o, the latest and the most advanced LLM to date that gives the state-of-the-art
quality of text generation and MT according to most users’ experience. The two MAS
outperform GPT-4o in all three quality dimensions and the ACS score.

In the dimension of accuracy in legal meaning, MAS 10 and 11 achieve an improvement
over GPT-4o by 4.60\% and 2.81\%, respectively. Interestingly, MAS 10 using GPT-3.5 Turbo as
Annotator agent outperforms MAS 11 using human error annotation by 1.75\%. This result suggests
that human annotation does not enable any better preservation of legal meaning in the translation
than automated annotation, although the former is still supposed to be superior to the latter.

In terms of coherence and cohesion in structure, GPT-4o still scores the lowest,
underperforming MAS 10 and 11 by 3.09\% and 3.43\%, respectively. Also, human error annotation
slightly outperforms automated annotation by 0.3\%, but whether this tiny margin indicates a
significant difference is without doubt questionable.

In the dimension of appropriateness in style, all three systems achieve extremely high
scores very close to the ceiling, indicating that all of them perform excellently in translation style
adaptability. Still, however, both MAS 10 and 11 outperform GPT-4o, one by 1.02\% and the other
1.43\%. In this dimension of evaluation, human annotations help MAS 11 achieve the highest S
score, outperforming MAS 10 with automated annotation slightly by 0.40\%.

According to the ACS score that unifies the three evaluation dimensions for each system,
MAS 10 gives the best overall performance and MAS 11 the second best, outperforming GPT-4o
by 3.87\% and 2.88\%, respectively. Given the overwhelmingly higher weight for meaning accuracy
in the ACS scoring, the overall effect of human annotation is not any better than that of using the
Annotator agent, in that the overall performance of MAS 11 is 0.85\% below that of MAS 10 in
terms of ACS scores.

In summary, these human evaluation results by the ACS scoring echo those by the two
automated evaluation metrics. In particular, MAS 10 of the highest configuration performs the best
in terms of overall translation quality in both evaluations, followed by MAS 11 of an as high
configuration except using human error annotation. Both of them outperform GPT-4o, the most
advanced LLM to date, in HK legal judgment translation.


\begin{table}[h!]
\centering
\label{fig:results_as_table2}
\caption{Results of human evaluation of the three representative MT systems, using the same labels A, S and C as in equation (1) for the three evaluation dimensions}
\begin{tabular}{lcccc}
\toprule
System & A & C & S & ACS \\
\midrule
GPT-4o & 8.91 & 9.05 & 9.82 & 9.04 \\
MAS 10 & \textbf{\underline{9.32 +4.60\%}} & 9.33 +3.09\% & 9.92 +1.02\% & \textbf{\underline{9.39 +3.87\%}} \\
MAS 11 & 9.16 +2.81\% & \textbf{\underline{9.36 +3.43\%}} & \textbf{\underline{9.96 +1.43\%}} & 9.30 +2.88\% \\
\bottomrule
\end{tabular}
\end{table}

\section{Cost Analysis}
The cost of human translation services can vary based on several factors, including the type
of text, the translator's location, and their level of experience. The American Translators
Association recommends a minimum charge of US\$0.12 per word for professional translation
services. Therefore, translating \textit{FACC 1/2021 [2021] HKCFA3} – a Final Criminal Appeal Case
decided by the Court of Final Appeal, which contains 11,585 English words, would cost
US\$1,390.20.

In contrast, the cost of translating the entire test set using GPT-4o is approximately
US\$0.39. Using the TAP, the cost for translating the entire test set breaks down to approximately
US\$0.08 (Translator) + US\$0.05 (Annotator) + US\$0.22 (Proofreader) = US\$0.35. Thus, using
the TAP to translate Hong Kong legal judgments can reduce translation costs by 3,972 times
compared to human translation and by 10.26\% compared to GPT-4o.\footnote{Note that US\$0.39 for using GPT-4o is an API cost, and US\$0.35 for using our multi-agent translator is also an API
cost. As the name suggests, “API cost” refers to the monetary expense associated with using an Application
Programming Interface. Such cost does not include the cost for using a human editor to proofread and edit the output
translation of an API. The average standard rate for human editing is approximately US\$0.04 per word (see e.g.,
\url{https://www.translationedge.com/pricing}). The editing cost for the said judgment would then be US\$0.04 x 11,585
words = US\$463.30. So the total cost for translating plus editing the judgment would be US\$0.35 + US\$463 =
US\$463.35, saving US\$926.85, or 3 times the full human translation cost.\label{fn:gpt4_cost}}

\section{Limitations and Future work}

\paragraph{The number of evaluated judgments is limited}
Due to time constraints, we have used
only one judgment for this paper as the evaluation set for the system proposed in this paper. If
conditions allow in the future, we will use a large-scale set of judgments for further evaluation.

\paragraph{Chunk-level editing functionality is limited}
During the final review and proofreading
stage based on human-computer interaction, we plan to add chunk replacement functionality,
including "replace entire text with one click" and "replace current chunk" options.

\paragraph{LLM's multi-turn dialogues exhibit hallucination}
When setting multiple rounds (3
rounds, 5 rounds) of dialogue between the Annotator LLM and the Proofreader LLM for repeated
revisions, we found that the meaning of the translation often deviates from the original text after
multiple revisions (hallucination phenomenon). The preliminary solution we propose, referencing
\cite{wu2024beyond}, is to add an extra hallucination arbitrator LLM. This part of
the work will be addressed in a subsequent paper.

\section{Summary}
In this second part, we have introduced the TAP, an innovative multi-agent virtual studio
designed to provide fast and high-quality translation services for Hong Kong legal judgments.
Through system design, implementation, and experimental validation, we have demonstrated that
this system effectively handles the inherent complexity and subtle nuances of legal judgments. By
incorporating three key roles—Translator, Annotator, and Proofreader—the system achieves a
highly coordinated translation and proofreading workflow. The specialized division of labor and
mutual collaboration among the agents significantly enhance translation performance.

Utilizing a few-shot prompting strategy embedded with translation/proofreading memory,
the system not only improves the performance of individual LLM agents but also enhances the
overall performance of the MAS in terms of translation and proofreading accuracy. This approach
ensures contextual coherence and terminological consistency.

We employed cutting-edge learned metrics—such as \texttt{XCOMET-XL} and \texttt{Wmt22-unite-da}
automatic evaluation metrics—as well as subjective fine-grained assessments by domain
translation experts with 30 years of experience, to comprehensively validate the exceptional
performance of the TAP system (outperforming the current state-of-the-art LLM, GPT-4o, across
multiple evaluation dimensions). Our empirical results (human evaluations) indicate that
translations generated by the TAP receive higher ratings from domain translation experts
compared to human-written reference translations, especially in terms of legal accuracy, stylistic
appropriateness, and structural coherence and consistency.

Lastly, we compared the translation costs between using the TAP system and GPT-4o,
finding that the former can save 3,972 times the cost. In summary, the TAP system exhibits
exceptional performance in handling complex legal judgment translation/proofreading tasks and
holds significant promise for widespread application. Looking ahead, we plan to further explore,
track, and integrate the most advanced large language models, optimizing the interactions and
collaboration strategies of the LLM agents within the system, with the aim of achieving even
greater breakthroughs in the field of Hong Kong legal judgment translation.

\newpage 
\ontribution{Our contributions}

\setcounter{section}{0}

\section{Construction of the CFA Judgement Corpus 97-22}
We have created a comprehensive bilingual (English-Chinese) legal judgment corpus for Hong Kong, spanning from 1997 to 2022, named the \textbf{CFA Judgement Corpus 97-22}. This open-source dataset is continuously maintained and available at: \href{https://huggingface.co/datasets/xxuan-nlp/CFA_Judgement_Corpus_97-22}{Hugging Face}.This resource is invaluable for legal researchers, translators, and professionals, providing a rich database for training and reference in legal translation and linguistic studies.

\section{Development of TAP}
We have pioneered the first multi-agent collaborative system for translating and proofreading Hong Kong legal judgments, named TAP. This system incorporates
an independent evaluation feedback module and integrates feedback mechanisms for selfcorrection in machine translation, forming an interpretable translation-proofreading system. This
innovation enhances the accuracy and reliability of legal translations, promoting greater trust and
efficiency in the bilingual legal system.

\section{Role of LLM Agents in the System}
\paragraph{Translator Agent:} The Translator agent is designed to automatically translate English judgments of Hong Kong into Traditional Chinese. This automation significantly speeds up the translation process, ensuring timely access to translated legal documents.

\paragraph{Error Annotation:} The Annotator agent utilizes a multi-level translation evaluation annotation standard (Proofread Codes) developed by experts in Hong Kong judgment translation. This agent performs domain-specific error annotation on the initial translations, ensuring high standards of accuracy and consistency.

\paragraph{Proofreader Agent:} The Proofreader agent conducts thorough domain-specific translation proofreading based on user inputs and the outputs of the Translator and Annotator agents. This final step ensures that the translations meet the required legal and linguistic standards, providing high-quality translated judgments.

\newpage 
\Impact{Long-Term Impacts}
These achievements mark a significant milestone in the journey towards a fully bilingual legal system in Hong Kong. By providing robust tools and resources for accurate and reliable legal translation, we are paving the way for enhanced accessibility and understanding of legal documents for all members of society, regardless of their language preference. Our work contributes to greater transparency and trust in the legal system, fostering a more inclusive and equitable legal environment in Hong Kong.

The platform's ability to incorporate vetted translations into its bilingual corpus as refined training data represents a significant step forward in machine learning. This iterative learning process means that the platform continuously improves its translation capabilities, producing higher quality translations over time. The use of the latest large language models and a multi-agent system ensures that the platform remains at the forefront of translation technology.

As the platform grows, the continuous input of translations from various sources will expand the corpus of case law in Chinese. This ever-growing body of translated judgments will evolve into a comprehensive and authoritative bilingual legal resource. This corpus will be invaluable for ongoing legal education, research, and practice, providing a rich database of legal precedents in both languages.

Moreover, incorporating English case law into the Chinese language provides a rich repository of judicial insights and reasoning techniques that can elevate the quality of Chinese judgments. As highlighted by the Supreme People’s Court, the essence of judgment writing lies in clear reasoning, logical rigor, and precise language. English case law, renowned for its comprehensive reasoning and thorough analysis, serves as an exemplary model in these aspects. By studying and integrating these judgments, Chinese judges can adopt and refine their own approaches to legal reasoning and writing, ensuring that their judgments are not only legally sound but also articulate and persuasive.

The translation project thus becomes a conduit for cross-jurisdictional learning and the exchange of judicial practices. It enables Chinese judges to gain direct exposure to the methodologies and analytical frameworks used by their counterparts in common law jurisdictions. This exposure can demystify the judicial processes of common law countries, providing practical examples of how complex legal issues are dissected and resolved through meticulous reasoning.

Furthermore, the translated judgments serve as a valuable educational resource for legal scholars, practitioners, and students in Hong Kong and mainland China. They offer concrete examples of high-caliber legal writing and reasoning, which can be incorporated into legal education and professional training programs. This ongoing exchange and learning process can help raise the overall standard of judicial decision-making, contributing to the development of a more robust and sophisticated legal system in Hong Kong.

In summary, translating English case law into Chinese is a multifaceted initiative that not only strengthens Hong Kong's bilingual legal system but also enriches the judicial practice by providing a wealth of judicial insights. This endeavor aligns with the Supreme People’s Court’s emphasis on clear and rigorous legal reasoning, ultimately promoting the development of a more transparent, accessible, and high-quality judiciary in China.

\newpage 
\way{The way forward}
While we strongly disagree with the Judiciary’s current policy on the translation of judgments, we support its “pragmatic approach” of addressing the immediate needs of judges, the legal profession, litigants, and the public at large. The HMIT platform we have developed, once made publicly accessible, will significantly contribute to meeting these pragmatic needs. For litigants intending to cite English precedents in trials conducted in Chinese, the platform offers a practical solution. By enabling them to translate and submit Chinese translations of precedents in advance, the platform aligns with the existing practice of vetting translations through the Court Translation Section of the Judiciary. This will ensure that all relevant legal texts are available in the appropriate language, maintaining the integrity and accuracy of legal proceedings.

However, our platform is only a prototype of a specialized system for judgment translation. A fully-fledged, comprehensive platform is essential—one that is supported by an extensive corpus encompassing various branches of law, equipped with advanced editing functions, and managed by a team of AI, translation, and legal professionals.

To achieve this, a committee comprising judges, legal scholars, and practicing lawyers should be established to prioritize and select the most frequently cited and legally significant judgments. This committee would implement a phased translation schedule, launch pilot translation projects for specific categories of judgments, and develop standardized translation protocols and best practices based on these projects. The Government, with its robust teams of IT experts in the Innovation, Technology and Industry Bureau, law translation officers in the Department of Justice, and court translators in the Judiciary, already possesses the necessary expertise. It needs only to allocate a small number of experts from these departments to oversee the platform’s operation.

Reinstating the Bilingual Laws Advisory Committee, composed of legal, translation, and language professionals, would provide the Judiciary with expert advice on judgment translation. In summary, the Government has all the expertise and resources required to implement a fully bilingual legal system. What is missing is the determination to execute it.

\newpage 

\renewcommand{\bibname}{References}

\bibliographystyle{unsrt}

\bibliography{references} 

\newpage 
\appendix
\renewcommand{\thetable}{\arabic{table}} 
\setcounter{table}{0} 

\section*{Appendix}

\renewcommand{\arraystretch}{1.3} 

\begin{table}[H]
\centering
\caption{Proofread Codes}
\begin{tabular}{|p{0.2\textwidth}|p{0.15\textwidth}|p{0.6\textwidth}|}
\hline
\textbf{Error Category} & \textbf{Subcategory} & \textbf{Description} \\
\hline
\multirow{10}{*}{\textbf{Accuracy}} 
 & CW & Choice of word. The word or expression is not a good choice. \\
\cline{2-3}
 & IF & Information structure not preserved. \\
\cline{2-3}
 & MC & Meaning has been changed because of inappropriate restructuring, e.g., changing the passive to active or vice versa. \\
\cline{2-3}
 & MT & Mistranslation due to inadequate comprehension or misinterpretation of the source text. \\
\cline{2-3}
 & NA & The translation conveys a different meaning from that of the source text. \\
\cline{2-3}
 & NC & Meaning not clear, e.g., because of ambiguity, vagueness or syntactic problems. \\
\cline{2-3}
 & OM & Omission. Part of the original has been left untranslated. \\
\cline{2-3}
 & OT & Over-translation. Too much has been read into the source text. \\
\cline{2-3}
 & TL & Too literal, affecting comprehensibility. \\
\cline{2-3}
 & UT & Under-translation. Meaning is not adequately captured in translation. \\
\hline
\multirow{12}{*}{\textbf{Grammar}} 
 & Art & Article. \\
\cline{2-3}
 & Det & Determiner. \\
\cline{2-3}
 & MD & Modality. \\
\cline{2-3}
 & NB & Number. \\
\cline{2-3}
 & PN & Punctuation. \\
\cline{2-3}
 & Prep & Wrong preposition. \\
\cline{2-3}
 & PS & Part of speech. \\
\cline{2-3}
 & SP & Spelling or wrong character. \\
\cline{2-3}
 & ST & The sentence or part of the sentence is ill-formed or ambiguous. \\
\cline{2-3}
 & SV & Subject verb agreement. \\
\cline{2-3}
 & TN & Tense problem. \\
\cline{2-3}
 & WO & Word order. \\
\hline
\multirow{9}{*}{\textbf{Usage and style}} 
 & CL & Collocation problem. \\
\cline{2-3}
 & CN & The word or expression has connotation not appropriate in the context. \\
\cline{2-3}
 & CO & Connective problem, e.g., inappropriate connectives. \\
\cline{2-3}
 & IC & Inconsistent use of a word; or incoherence between clauses or sentences. \\
\cline{2-3}
 & ID & Idiomaticity, i.e., unidiomatic expression. \\
\cline{2-3}
 & RF & Reference problem, e.g., ambiguous use of a pronoun. \\
\cline{2-3}
 & RN & Redundancy: the word or expression should be deleted. \\
\cline{2-3}
 & SL & Stylistic problems, e.g., the word or expression is not of an appropriate style. \\
\cline{2-3}
 & TS & Transition problems: sentences not well connected; bad language flow. \\
\hline
\end{tabular}
\end{table}

\begin{table}[H]
\centering
\caption{Zero-shot Translator Prompt. \texttt{src}: Original judgment text; Red text: Translator task prompt.}
\label{tab:AT2}
\begin{tabular}{@{}p{\textwidth}@{}}
\toprule
\textbf{Zero-shot Translator} \\ \midrule
\texttt{[Translator Role Prompt]} \\ \addlinespace
\textbf{\textcolor{lightred}{\texttt{English text: \{src\}}}} \\ \addlinespace
\textbf{\textcolor{lightred}{\texttt{Translate to Traditional Chinese text:}}} \\ \bottomrule
\end{tabular}
\end{table}

\begin{table}[H]
\centering
\caption{Few-shot Translator Prompt. \texttt{\{src\}}: Original judgment text; \texttt{\{example\_src\_i\}}: the original judgment text of example i; \texttt{\{example\_tar\_i\}}: the initial translation text of example i; Blue text: Translator example prompt; Red text: Translator task prompt.}
\label{tab:AT3}
\begin{tabular}{@{}p{\textwidth}@{}}
\toprule
\textbf{Few-shot Translator} \\ \midrule
\texttt{[Translator Role Prompt]} \\ \addlinespace
\textbf{\textcolor{lightblue}{\texttt{English text: \{example\_src\_1\}}}} \\
\textbf{\textcolor{lightblue}{\texttt{Translate to Traditional Chinese text: \{example\_tar\_1\}}}} \\ \addlinespace
\textbf{\textcolor{lightblue}{\texttt{English text: \{example\_src\_2\}}}} \\
\textbf{\textcolor{lightblue}{\texttt{Translate to Traditional Chinese text: \{example\_tar\_2\}}}} \\ \addlinespace
\textbf{\textcolor{lightblue}{\texttt{English text: \{example\_src\_3\}}}} \\
\textbf{\textcolor{lightblue}{\texttt{Translate to Traditional Chinese text: \{example\_tar\_3\}}}} \\ \addlinespace
\textbf{\textcolor{lightblue}{\texttt{English text: \{example\_src\_4\}}}} \\
\textbf{\textcolor{lightblue}{\texttt{Translate to Traditional Chinese text: \{example\_tar\_4\}}}} \\ \addlinespace
\textbf{\textcolor{lightblue}{\texttt{English text: \{example\_src\_5\}}}} \\
\textbf{\textcolor{lightblue}{\texttt{Translate to Traditional Chinese text: \{example\_tar\_5\}}}} \\ \addlinespace
\textbf{\textcolor{lightred}{\texttt{English text: \{src\}}}} \\
\textbf{\textcolor{lightred}{\texttt{Translate to Traditional Chinese text:}}} \\ \bottomrule
\end{tabular}
\end{table}

\begin{table}[H]
\centering
\caption{Annotator Prompt. \texttt{\{src\}}: Original judgment text; \texttt{\{tar\}}: Initial translation text; Red text: Annotator task prompt.}
\label{tab:AT4}
\begin{tabular}{@{}p{\textwidth}@{}}
\toprule
\textbf{Annotator} \\ \midrule
\texttt{[Annotator Role Prompt]} \\ \addlinespace
\textbf{\textcolor{lightred}{\texttt{Source text: \{src\}}}} \\ \addlinespace
\textbf{\textcolor{lightred}{\texttt{machine translation: \{tar\}}}} \\ \addlinespace
\textbf{\textcolor{lightred}{\texttt{Annotated errors: (Do not output in separate lines; output only in one line.)}}} \\ \bottomrule
\end{tabular}
\end{table}

\begin{table}[H]
\centering
\caption{Zero-shot Proofreader Prompt. \texttt{\{src\}}: Original judgment text; \texttt{\{tar\}}: Initial translation text; \texttt{\{Feedback\_Annotated\_errors\}}: The output of the Annotator Agent. Red text: Proofreader task prompt.}
\label{tab:AT5}
\begin{tabular}{@{}p{\textwidth}@{}}
\toprule
\textbf{Zero-shot Proofreader} \\ \midrule
\texttt{[Proofreader Role Prompt Content]} \\ \addlinespace
\textbf{\textcolor{lightred}{\texttt{Source text: \{src\}}}} \\ \addlinespace
\textbf{\textcolor{lightred}{\texttt{machine translation: \{tar\}}}} \\ \addlinespace
\textbf{\textcolor{lightred}{\texttt{Annotated errors: \{Feedback\_Annotated\_errors\}}}} \\ \addlinespace
\parbox{\textwidth}{\textbf{\textcolor{lightred}{\texttt{Final translation: (Do not output in separate lines; output only in one line.)}}}} \\ \bottomrule
\end{tabular}
\end{table}

\begin{table}[H]
\centering
\caption{Few-shot Proofreader Prompt. \texttt{\{src\}}: Original judgment text; \texttt{\{example\_trans\_src\_i\}}: the original judgment text of example i; \texttt{\{example\_Feedback\_Annotated\_errors\_i\}}: the feedback\_Annotated\_errors of example i; \texttt{\{example\_prof\_tar\_i\}}: the final translation text of example i. \texttt{\{tar\}}: Initial translation text; \texttt{\{example\_Feedback\_Annotated\_errors\}}: The output of the Annotator Agent; Blue text: Proofreader example prompt; Red text: Proofreader task prompt.}
\label{tab:AT6}
\begin{tabular}{@{}p{\textwidth}@{}}
\toprule
\textbf{Few-shot Proofreader} \\ \midrule
\texttt{[Proofreader Role Prompt Content]} \\ \addlinespace
\textcolor{lightblue}{\texttt{Source text: \{example\_trans\_src\_1\}}} \\
\textcolor{lightblue}{\texttt{machine translation: \{example\_prof\_src\_1\}}} \\
\textcolor{lightblue}{\texttt{Annotated errors: \{example\_Feedback\_Annotated\_errors\_1\}}} \\
\textcolor{lightblue}{\texttt{Final translation: \{example\_prof\_tar\_1\}}} \\ \addlinespace
\textcolor{lightblue}{\texttt{Source text: \{example\_trans\_src\_2\}}} \\
\textcolor{lightblue}{\texttt{machine translation: \{example\_prof\_src\_2\}}} \\
\textcolor{lightblue}{\texttt{Annotated errors: \{example\_Feedback\_Annotated\_errors\_2\}}} \\
\textcolor{lightblue}{\texttt{Final translation: \{example\_prof\_tar\_2\}}} \\ \addlinespace
\textcolor{lightblue}{\texttt{Source text: \{example\_trans\_src\_3\}}} \\
\textcolor{lightblue}{\texttt{machine translation: \{example\_prof\_src\_3\}}} \\
\textcolor{lightblue}{\texttt{Annotated errors: \{example\_Feedback\_Annotated\_errors\_3\}}} \\
\textcolor{lightblue}{\texttt{Final translation: \{example\_prof\_tar\_3\}}} \\ \addlinespace
\textcolor{lightblue}{\texttt{Source text: \{example\_trans\_src\_4\}}} \\
\textcolor{lightblue}{\texttt{machine translation: \{example\_prof\_src\_4\}}} \\
\textcolor{lightblue}{\texttt{Annotated errors: \{example\_Feedback\_Annotated\_errors\_4\}}} \\
\textcolor{lightblue}{\texttt{Final translation: \{example\_prof\_tar\_4\}}} \\ \addlinespace
\textcolor{lightblue}{\texttt{Source text: \{example\_trans\_src\_5\}}} \\
\textcolor{lightblue}{\texttt{machine translation: \{example\_prof\_src\_5\}}} \\
\textcolor{lightblue}{\texttt{Annotated errors: \{example\_Feedback\_Annotated\_errors\_5\}}} \\
\textcolor{lightblue}{\texttt{Final translation: \{example\_prof\_tar\_5\}}} \\ \addlinespace
\textcolor{lightred}{\texttt{Source text: \{src\}}} \\
\textcolor{lightred}{\texttt{machine translation: \{tar\}}} \\
\textcolor{lightred}{\texttt{Annotated errors: \{Feedback\_Annotated\_errors\}}} \\
\textcolor{lightred}{\texttt{Final translation:}} \\ \bottomrule
\end{tabular}
\end{table}

\end{document}